\documentclass[11pt, hidelinks]{article}

\usepackage[utf8]{inputenc}
\usepackage[font=small,labelfont=bf]{caption}
\usepackage{authblk}
\usepackage{orcidlink}
\usepackage{microtype}
\usepackage{hyperref}
\usepackage[style=ieee]{biblatex}
\usepackage{enumitem}
\usepackage{booktabs}
\usepackage{tabularx}
\usepackage{float}
\usepackage{ragged2e}
\usepackage{amsmath}
\usepackage{array}
\usepackage{multirow}
\usepackage{caption}

\newcolumntype{C}{>{\Centering\arraybackslash}X}

\setlist[itemize]{noitemsep, topsep=0pt}

\title{\textbf{A Data Efficiency Study of Synthetic Fog for Object Detection Using the Clear2Fog Pipeline}}

\author{Mohamed Ahmed Mohamed \orcidlink{0009-0007-9455-1346}\thanks{Corresponding author}}
\author{Xiaowei Huang \orcidlink{0000-0001-6267-0366}}
\affil{School of Computer Science and Informatics, University of Liverpool, Liverpool, L69 3DR, UK \\ {\small \texttt{\{M.A.Mohamed2, xiaowei.huang\}@liverpool.ac.uk}}}

\date{}

\begin{document}

\maketitle

\begin{abstract}
\noindent Object detection in adverse weather is critical for the safety of autonomous vehicles; however, the scarcity of labelled, real-world foggy data remains a significant bottleneck. In this paper, we propose Clear2Fog (C2F), an end-to-end, physics-based pipeline that simulates fog for clear-weather datasets under a unified camera and LiDAR framework. C2F combines monocular depth estimation with a novel atmospheric light estimation method to improve the physical consistency of synthetic fog generation while reducing structural artefacts and chromatic biases observed in existing frameworks. Utilising a training set of 270,000 images from the Waymo Open Dataset, we conduct an extensive data efficiency study to investigate whether environmental diversity can reduce dataset scale requirements and improve model generalisation under varying fog conditions. Our findings reveal that models trained on mixed-density fog datasets at 75\% scale achieve performance that is not statistically different from those trained on fixed-density datasets at 100\% scale, reducing synthetic training data requirements by 25\%. We observe that this efficiency trend is consistent across two representative detector architectures. Furthermore, we investigate the sim-to-real transfer by using C2F-generated data as a pre-training foundation before fine-tuning on real-world fog data. We demonstrate that, within the evaluated settings, a relative 10x increase in the default fine-tuning learning rate reduces the negative transfer caused by standard fine-tuning, resulting in an absolute increase of up to 0.0117 mAP beyond the real-only baseline. Overall, this work demonstrates the value of diverse synthetic fog as a pre-training tool for real-world adaptation. The source code for the pipeline and all the experimental configurations are available at: \url{https://github.com/mmohamed28/Clear2Fog}. 

\vspace{0.5cm} 
\end{abstract}

\bigskip
\section{Introduction}
Autonomous vehicles (AVs) have gained significant attention in recent years due to their potential, convenience, efficiency and economic benefits. The perception system in AVs transforms sensory data into semantic information~\cite{r1}; however, adverse weather conditions such as fog, snow and rain pose significant challenges for these systems. As AV perception is fundamental to its navigation, ensuring its robustness in such conditions is critical. The degradation of an AV’s perception system in fog is caused by the scattering and absorption of light as it travels through the atmosphere~\cite{r2}. The suspended particles affect visibility and reduce focus by causing a loss of colour and feature information of objects within a scene, which in turn affects the performance of deep learning models and impacts the perception of scene depth~\cite{r3, r4}. Furthermore, these particles reduce image contrast by deflecting and diffusing light rays, affecting the recognition of patterns and edges~\cite{r3, r4}. While the impact on perception through cameras is primarily visual, fog also affects LiDAR sensors by weakening and scattering laser signals passing through the atmosphere~\cite{r5}. This leads to the false detection of objects and the distortion of their perceived shape and position in the scene.

The main bottleneck for developing robust perception in foggy conditions is the lack of large-scale datasets that are suitable for training and evaluating modern detection models. Creating real foggy datasets presents many challenges; they are dependent on unpredictable weather patterns and require significant time and financial resources. Furthermore, capturing high-density traffic scenes in fog is difficult as these conditions typically lead to fewer vehicles and pedestrians on the streets, resulting in fewer objects per frame compared to clear-weather data. Current publicly available datasets that contain real-world foggy scenes are rare, and while datasets like Seeing Through Fog (STF)~\cite{r6} exist, they are limited in scale. To overcome this, many researchers have generated synthetic foggy datasets from standard clear-weather data~\cite{r7, r8, r9, r10} or have combined subsets of multiple foggy datasets to increase data scale and diversity~\cite{r11, r12, r13}. However, most current efforts are static, task-specific and lack a unified approach for generating consistent fog across both camera and LiDAR data. Moreover, while the benefits of data scale are well-documented for clear-weather conditions~\cite{r14, r15}, the data efficiency of synthetic adverse weather remains significantly underexplored, particularly how dataset scale and environmental diversity contribute to model robustness.

\begin{figure}[t]
    \centering
    \includegraphics[width=\textwidth]{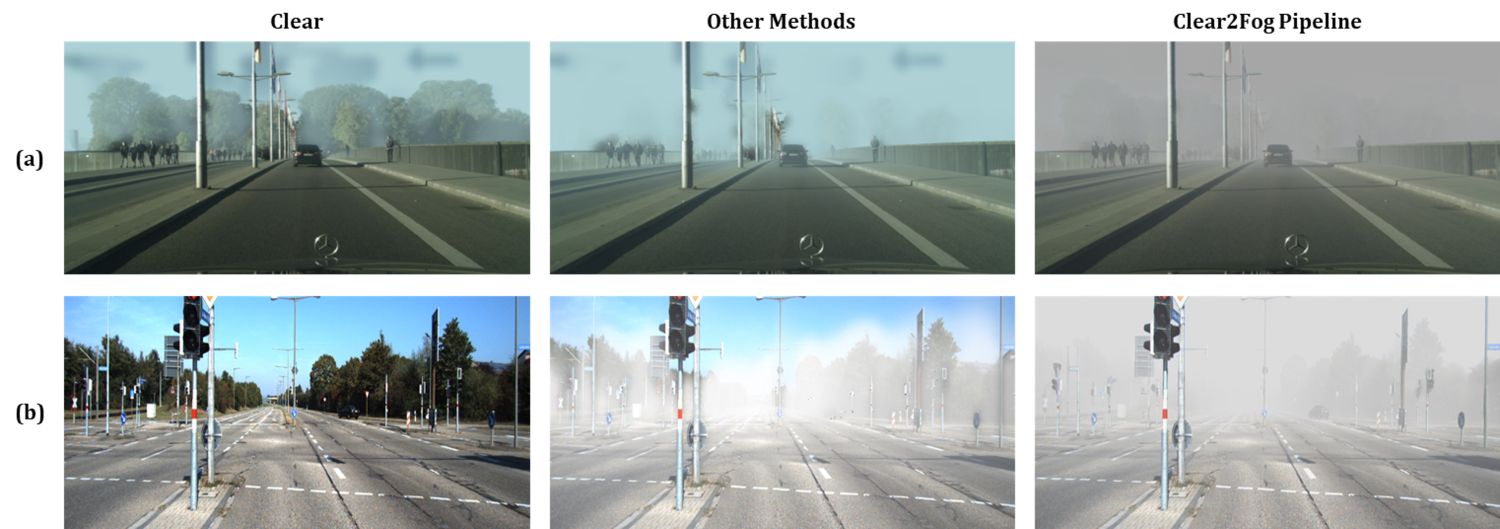}
    \caption{Qualitative comparison of fog simulation realism between the proposed Clear2Fog (C2F) pipeline and other established methods ((a) Foggy Cityscapes~\cite{r16} and (b) Multifog KITTI~\cite{r7}). (a) Illustrates the removal of chromatic bias in C2F using a luminance-clipping method for a physically grounded, colour-neutral output as opposed to unnatural colour casts introduced in the established methods. (b) Demonstrates scene consistency by ensuring consistent atmospheric occlusion across the entire scene in C2F.}
    \label{fig1}
\end{figure}

To address these challenges, this paper introduces Clear2Fog (C2F), an end-to-end, physics-based pipeline for generating large-scale multimodal foggy datasets. The pipeline incorporates atmospheric scattering models to simulate fog across both camera and LiDAR data under a unified framework. It integrates a monocular metric depth estimation model to improve fog synthesis in regions beyond the range of traditional LiDAR sensors, such as distant areas and sky regions. As shown in Figure~\ref{fig1}, C2F addresses two key limitations observed in existing fog simulation methods. Firstly, it improves depth representation by using monocular depth estimation rather than relying on sparse LiDAR depth completion to allow distant regions and the sky to be treated as beyond the effective visibility range during fog simulation. Secondly, it improves physical realism by introducing a novel atmospheric light estimation method based on luminance-clipping grounded in an empirical analysis of over 2,000 real-world foggy images. This reduces the chromatic biases (e.g. blue or green tints) observed in existing simulation methods and produces atmospheric scattering more consistent with Mie scattering principles.

Utilising the C2F pipeline, we conduct a comprehensive data efficiency study to investigate how dataset size and environmental diversity influence object detection performance by leveraging a training set of 270,000 images from the Waymo Open Dataset~\cite{r14}. We evaluate the impact of synthetic fog by training models with diverse dataset scales and densities to determine how these factors influence synthetic pre-training for adverse weather perception. To assess the perceptual realism of our pipeline, we conduct a human perceptual study involving 440 pairwise judgements, where C2F-generated images are preferred 92.95\% of the time over an established method. Quantitatively, our results reveal that models trained on a mixed-density fog distribution at 75\% scale achieve performance that is not statistically different from those trained on fixed-density models at 100\% scale, reducing synthetic training data requirements by 25\%. Finally, we investigate the sim-to-real transfer by using synthetic data as a pre-training foundation for real-world adaptation. We find that standard fine-tuning rates cause negative transfer and allow models to retain synthetic biases that degrade real-world performance. However, a controlled learning rate sensitivity analysis shows that a 10$\times$ increase in the default fine-tuning learning rate provides the highest adaptation performance within the evaluated range, leading to reduced negative transfer. This strategy reduces the negative transfer observed during standard fine-tuning and yields an absolute increase of up to 0.0117 mAP beyond the real-only baseline. 

\medskip
\noindent The primary contributions of this paper are as follows:
\begin{itemize}
    \item We present Clear2Fog (C2F), an end-to-end, physics-based pipeline that generates consistent fog across camera and LiDAR modalities for clear-weather datasets. The pipeline addresses the issue of chromatic biases and preserves scene structure through the integration of monocular metric depth estimation and a novel, colour-neutral atmospheric light estimation method.
    \item We conduct a human perceptual study as a supporting qualitative indicator of the pipeline’s realism across 440 pairwise judgements, with C2F-generated images being preferred 92.95\% of the time over an established framework. 
    \item We demonstrate that environmental diversity is an important factor in synthetic data efficiency. Specifically, while increasing dataset scale improves detection performance, models trained on a 75\% scale mixed-density fog dataset achieve performance that is not statistically different from those trained on a 100\% scale fixed-density dataset. This trend is consistent across representative two-stage and one-stage detector architectures (Faster R-CNN and YOLOX-S).
    \item We investigate sim-to-real adaptation through synthetic pre-training and real-world fine-tuning using a controlled learning rate sensitivity analysis. We show that a relative tenfold increase in the default fine-tuning learning rate reduces the negative transfer observed during standard fine-tuning. This achieves an absolute improvement of up to 0.0117 mAP beyond the real-only baseline.
\end{itemize}

\medskip
\noindent The remainder of the paper is organised as follows. Section~\ref{sec:related_works} reviews the related work. Section~\ref{sec:c2f_pipeline} details the methodology and implementation of the Clear2Fog pipeline. Section~\ref{sec:experiments} describes the experimental evaluation and analyses the results. Finally, Section~\ref{sec:conclusion} concludes the paper and discusses potential directions for future research.

\bigskip
\section{Related Works}
\label{sec:related_works}

\subsection{Simulated Foggy Datasets}
While several high-impact autonomous driving datasets exist, such as nuScenes \cite{r15} and KITTI~\cite{r17}, this study utilises the Waymo Open Dataset~\cite{r14} due to its superior scale and annotation quality. Historically, the KITTI dataset has been a popular foundation for fog simulation. For instance, Mai et al.~\cite{r7} developed Multifog KITTI by applying physics-based fog simulation across the entire KITTI dataset, including the stereo camera and LiDAR modalities. Similarly, Oh et al.~\cite{r8} created Foggy KITTI using the camera images only, whereas Wang et al.~\cite{r9} utilised a hybrid approach by combining the original dataset with weather-augmented versions. Expanding beyond KITTI, Wu et al.~\cite{r10} addressed data scarcity by introducing Fog-nuScenes and combining it with the original nuScenes dataset. These approaches leverage the rich annotations and diverse scenarios of established datasets.

Another strategy to address foggy data scarcity is aggregating images from multiple sources to create a unified dataset with a wider variety of scenes. For instance, Patel et al.~\cite{r11} created the Urban Weather Diversity Dataset (UWDD) by combining 3,000 images from KITTI, Udacity~\cite{r18} and the Indian Driving Dataset (IDD)~\cite{r19}. Similarly, He and Liu~\cite{r12} constructed a foggy dataset by capturing 1000 images from two cities in China and combining them with foggy images from BDD100K~\cite{r20}, Oxford RobotCar~\cite{r21} and Apolloscape~\cite{r22}. Recently, Shen et al.~\cite{r13} utilised images from Oxford RobotCar, nuScenes and DrivingStereo~\cite{r23} to test their monocular depth estimation model in foggy conditions.

Despite these efforts, current methods face some limitations in terms of their physical realism and scalability. While these custom datasets are valuable for testing specific models, they are often small-scale, task-specific and not always reproducible. This shows that while the research community has recognised the issue of foggy data scarcity, a general-purpose and scalable solution is yet to be established. Furthermore, many established simulations rely on sensor-based depth completion, which is restricted by the sensor’s range and often leaves distant regions entirely clear. They also frequently introduce unnatural chromatic biases due to heuristic-based atmospheric light estimation. 

\subsection{Camera-Based Fog Simulation}
Simulating fog on camera images has mainly followed two directions: physics-based methods and learning-based generative methods. Physics-based methods build upon the standard optical model of Koschmieder’s law~\cite{r24}, which models the process of light attenuation and the addition of atmospheric light. Sakaridis et al.~\cite{r16} adapted this model for autonomous driving by applying it to the Cityscapes dataset~\cite{r25} to create Foggy Cityscapes, which is an established method for evaluating semantic segmentation under adverse conditions. Bernuth et al.~\cite{r26} argued that fog affects the three RGB channels differently and assigned different extinction coefficients for each channel. Sen et al.~\cite{r27} and Zhang et al.~\cite{r28} proposed the use of Perlin noise to introduce spatial randomness to reflect real-world heterogeneity. Additionally, Zhang et al.~\cite{r28} proposed estimating atmospheric light by randomly sampling from a pre-collected database of sky luminance vectors derived from 500 real foggy images. However, this strategy lacks the dynamic robustness provided by direct image-based estimation and can introduce chromatic biases if the selected database vector does not match the original scene’s lighting context.

Other physics-based methods have focused on increasing environmental complexity. The FoHIS method~\cite{r29} simulates heterogeneous fog by applying 3D Perlin noise to the attenuation coefficient and modelling the effect of elevation on fog density. Beregi-Kovacs et al.~\cite{r30} proposed a physics-based algorithm based on the Radiative Transfer Equation (RTE) to model anisotropic scattering. While this is a physically comprehensive method, the use of large angular and spatial tensors makes it computationally expensive as it requires large memory and long inference times compared to the closed-form nature of the Koschmieder model.

Alternatively, learning-based approaches utilise Generative Adversarial Networks (GANs)~\cite{r31} for weather domain translation. Due to the scarcity of paired clear and foggy images, unpaired image-to-image translation is typically preferred~\cite{r32}. For instance, Li et al.~\cite{r33} developed a weather GAN capable of manipulating specific weather cues to transform the weather conditions in an image while preserving the irrelevant areas. Musat et al.~\cite{r34} further proposed a unified generator architecture for multi–weather augmentation across seven different conditions. More recent efforts have expanded this domain translation paradigm into layout-to-image perception loops that leverage dual-task cycle-consistency frameworks~\cite{r61} and visually prompted bias recalibration synthesis engines~\cite{r62}, which look to optimise downstream model robustness. Other hybrid approaches have explored modern rendering engines such as DigiWeather~\cite{r35} or the inversion of dehazing networks like GridNet~\cite{r36}. Nevertheless, these methods are often limited in their generalisability or are restricted to single-viewpoint datasets, which limit their application to large-scale, multimodal autonomous driving datasets.

In this work, we adopt a physics-based approach due to its controllability, generalisability and its relative computational efficiency. Furthermore, we improve upon the approach in~\cite{r28} by replacing the static 500-vector database with a luminance-clipping process grounded in an empirical study of over 2,000 real-world images. This produces colour neutrality that is consistent with real-world fog and provides dynamic estimation through direct image-based calculation. We also address the structural failures inherent in LiDAR-based depth completion by integrating a monocular metric depth estimation model to improve the physical consistency of the entire scene.

\subsection{LiDAR-Based Fog Simulation}
As opposed to camera-based methods, there isn’t an established standard in the literature for simulating fog on LiDAR point clouds. Current research is categorised into physics-based, probabilistic and learning-based methods. Rasshofer et al.~\cite{r37} established a physics-based model derived from the optical physics of LiDAR sensors where the received signal is the convolution of the transmitted power and the spatial impulse response of the environment. In foggy conditions, this is modelled as the sum of a hard target response (i.e. attenuation from solid objects) and a soft target response (i.e. backscattering from fog particles). Hahner et al.~\cite{r38} adopted this framework to provide a simple algorithm to simulate fog on any clear-weather point cloud. Although computationally intensive, this method’s modelling of soft targets provides a physically more accurate method compared to simple attenuation heuristics.

Bijelic et al.~\cite{r6} proposed another physics-based model using a first-order approximation of Koschmieder’s Law for active sensors that focuses on attenuation and includes a noise-floor threshold. While computationally lightweight, this model was designed to reproduce measurements carried out in a 30-metre fog chamber. Other complex hybrid methods, like LISA~\cite{r39} and the virtual LiDAR model by Haider et al.~\cite{r40}, employ Monte-Carlo simulations and Mie scattering theory to account for optical losses and inherent detector noise. However, these methods are often difficult to reproduce and less accessible for large-scale, general-purpose simulation pipelines.

Alternative techniques include probabilistic and data-driven approaches. Teufel et al.~\cite{r41} proposed a probabilistic model that uses exponential functions to determine the likelihood of a point being deleted (i.e. attenuation) or moved towards the sensor (i.e. backscatter). While efficient, such methods lack the reproducibility of deterministic optical models without random seed controls. Similarly, learning-based efforts have utilised CycleGAN architectures~\cite{r32, r42} or two-stage frameworks like LaNoising~\cite{r43}, which uses Gaussian Process Regression to predict detection ranges. More recently, LiDARWeather~\cite{r44} combined selective jittering with a Deep Q-Network for point removal. Despite their potential for realism, these methods often suffer from high errors in distance and intensity or require extensive domain-specific training.

Among the surveyed techniques, physics-based models stand out as the most interpretable and controllable approach to simulating fog on LiDAR point clouds. In this work, we adopt the method by Hahner et al.~\cite{r38} as it combines a solid theoretical foundation with a practical implementation that integrates smoothly into fog simulation pipelines. We extend this by ensuring multimodal synchronisation between camera and LiDAR simulations and provide a scalable solution that provides consistency across the entire dataset.

\subsection{Importance of Dataset Scale}
It is widely established in literature that deep learning models are “data hungry”, requiring large amounts of data to train effectively. Studies such as those by Kaplan et al.~\cite{r45} and Sun et al.~\cite{r46} have demonstrated that model performance scales predictably with increases in both dataset size and model capacity. Since autonomous vehicles mainly rely on deep learning algorithms for information extraction (e.g. object detection) and decision making, large amounts of data are needed to improve awareness of their surroundings~\cite{r47}.

This concept has been demonstrated in practice with major benchmark datasets. Both Caesar et el.~\cite{r15} and Sun et al.~\cite{r14} have showed that model performance improves directly as the percentage of training data used increases. As noted in~\cite{r15}, the full potential of complex architectures can only be verified through larger and more diverse training sets. 

While the benefits of data scale under clear conditions are well-documented, the impact of scale within the adverse environmental domains (e.g. fog) remains largely underexplored. Mai et al.~\cite{r7} highlight that while large-scale labelled data produce the best results, the challenge of acquiring such data in foggy conditions remain a major bottleneck. We address this gap by utilising the Clear2Fog (C2F) pipeline to conduct a systematic data efficiency study. Unlike traditional scaling analyses that focus primarily on size, we investigate the relationship between data scale and environmental diversity in affecting model performance.

\bigskip
\section{Clear2Fog Pipeline}
\label{sec:c2f_pipeline}

\subsection{Pipeline Architecture and Methodological Novelty}

\subsubsection{Pipeline Architecture Overview}
\begin{figure}[t]
    \centering
    \includegraphics[width=\textwidth]{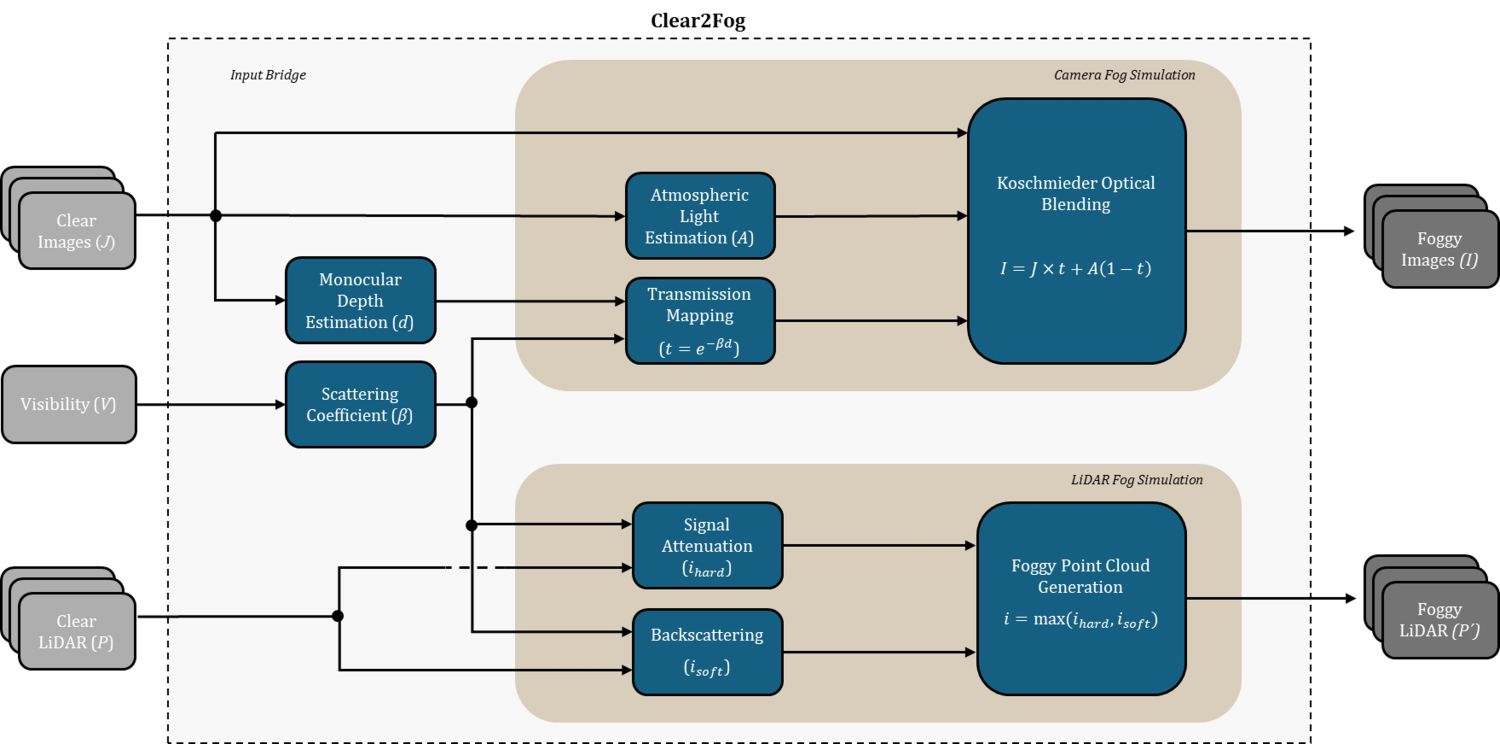}
    \caption{High-level architecture of the Clear2Fog pipeline.}
    \label{fig2}
\end{figure}

The Clear2Fog (C2F) pipeline is an end-to-end framework designed to generate consistent and configurable fog on clear-weather multimodal data. As shown in Figure \ref{fig2}, the pipeline takes in clear-weather RGB images ($J$), clear-weather LiDAR point clouds ($P$) and a target visibility parameter ($V$) to generate synchronised foggy outputs. The pipeline architecture is structured around three main pillars.

\textbf{The Input Bridge.} This stage prepares the data for simulation. The raw clear-weather images are processed by a monocular depth estimator to create a dense depth map ($d$) that provides the 3D spatial context for the scene. At the same time, the visibility parameter ($V$) is translated into a scattering coefficient ($\beta$), which acts as the unified control variable for the whole pipeline.

\textbf{Camera Fog Simulation.} This module utilises a physics-based scattering model grounded in Koschmieder’s law~\cite{r24}. It estimates the atmospheric light ($A$) through a luminance-clipping process and computes a transmission map ($t$) derived from the scene depth ($d$) and scattering coefficient ($\beta$). These components are combined in the Koschmieder Optical Blending stage to produce the final foggy image ($I$).

\textbf{LiDAR Fog Simulation.} This module simulates the degradation of point clouds through signal attenuation and backscattering. Using the shared scattering coefficient ($\beta$), it calculates both the attenuated hard target return ($i_{hard}$) and the soft target return ($i_{soft}$). The Foggy Point Cloud Generation stage performs a max-intensity selection, modelling how a LiDAR sensor perceives either a solid object or a dense patch of fog.

\subsubsection{Pipeline Contributions}
To clarify the distinction between the integration of established methods and the main contributions of the Clear2Fog (C2F) pipeline, Table~\ref{tab_methods} summarises our technical contributions against prior frameworks. While C2F utilises foundational components, including Koschmieder’s law~\cite{r24}, LiDAR scattering principles defined by Hahner et al.~\cite{r38} and Depth Pro~\cite{r49}, its contribution lies in providing a generalised and dynamically configurable framework. Previous methods, like ~\cite{r7, r16}, typically provide pre-generated datasets restricted to specific sensors or geographical domains; however, C2F provides an end-to-end pipeline that allows users to synthesise fog on arbitrary clear-weather datasets. This unified structure allows visibility levels to be configured across camera, LiDAR or joint sensor modalities depending on the available data channels.

Furthermore, C2F introduces a novel atmospheric light ($A$) estimation method consisting of two stages. Firstly, the framework extends the Dark Channel Prior method by Tang et al.~\cite{r53} by applying a 1,000m depth mask to restrict atmospheric light estimation to pixels corresponding to distant out-of-range regions (e.g. sky areas). Secondly, an empirically derived luminance-clipping method restricts the atmospheric light vector to a physically grounded daytime fog range derived from real-world data. Together, these techniques promote spectral neutrality consistent with Mie scattering physics and reduce the chromatic biases observed in prior fog simulation methods.

\begin{table}[p]
\centering
\caption{Contributions of the Clear2Fog (C2F) pipeline against established fog simulation baselines.}
\label{tab_methods}
\small
\renewcommand{\arraystretch}{2.0}

\begin{tabularx}{\textwidth}{C C C C}
\toprule
\textbf{Pipeline Property} & \textbf{Established Baselines} & \textbf{Proposed C2F Pipeline} & \textbf{Impact} \\ \midrule

Generalisable Framework & 
Static, pre-generated datasets restricted to specific sensors and geographical domains & 
Dataset-agnostic, open-source pipeline applicable to diverse user-defined clear-weather datasets (e.g. Waymo~\cite{r14}, COCO~\cite{r56}, Flickr30k~\cite{r57}) & 
Enables flexible synthetic fog generation across different autonomous driving datasets \\ 

Sensor Configuration & 
Limited to specific sensor configurations or single-sensor applications & 
Dynamically selectable framework (camera-only, LiDAR-only or joint) & 
Supports flexible compatibility across camera and LiDAR sensor configurations depending on user-provided channels \\ 

Atmospheric Light ($A$) strategy & 
Unconstrained dark channel sampling or static database vectors & 
Depth-constrained dark channel sampling with empirical luminance-clipping & 
Improves atmospheric colour consistency and reduces chromatic biases in generated fog \\  \bottomrule
\end{tabularx}
\end{table}

\subsection{Depth Estimation}
An important prerequisite for physics-based fog simulation is a per-pixel metric depth map ($d$), which provides the spatial foundation for the atmospheric scattering model. Although the pipeline can receive sparse LiDAR data ($P$), we found that traditional depth completion methods that densify sparse point clouds are less suitable for realistic fog simulation as they remain limited by the spatial coverage of the original measurements. Furthermore, since LiDAR data is not always available across different datasets, we utilise monocular depth estimation to provide a more universally compatible simulation pipeline.

\subsubsection{Model Selection and Scene Consistency}
Preliminary qualitative evaluations revealed that the evaluated depth completion model, Marigold-DC~\cite{r48}, produced inaccurate depth estimates in regions beyond the LiDAR sensor’s range, particularly in sky regions. As optical fog models are highly sensitive to depth, these errors can introduce structural inconsistencies where distant regions remain clear while the foreground is correctly degraded.

\begin{figure}[t]
    \centering
    \includegraphics[width=\textwidth]{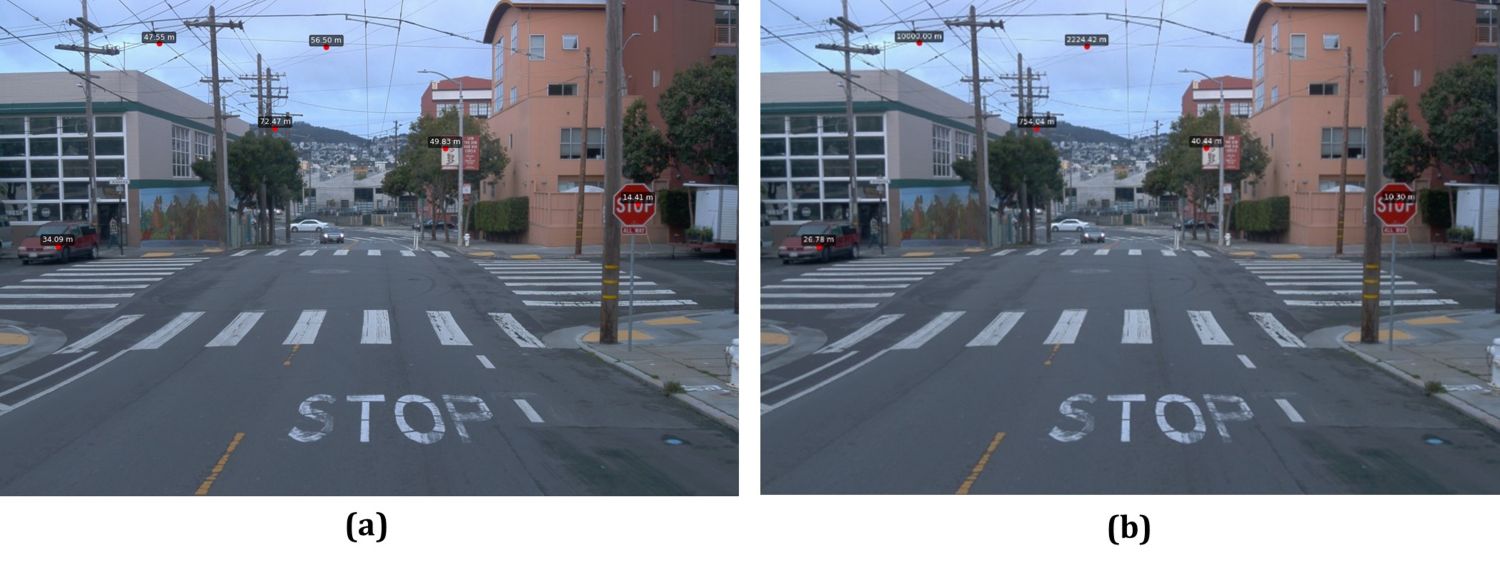}
    \caption{Qualitative comparison between two depth models. (a) Depth completion model via Marigold-DC~\cite{r48}. (b) Monocular depth estimation model via Depth Pro~\cite{r49}. The frame is from the Waymo Open Dataset~\cite{r14}.}
    \label{fig3}
\end{figure}

To preserve scene consistency, the C2F pipeline utilises a monocular depth estimation approach through the Depth Pro model~\cite{r49}. As illustrated in Figure~\ref{fig3}, while both depth methods perform adequately on foreground objects (e.g. signs and nearby vehicle), the depth completion approach produces physically unreasonable depth values in the sky region (e.g. 40m-50m). By estimating depth directly from the RGB image without depending on sparse LiDAR points, Depth Pro provides a more consistent representation of distant and sky regions where LiDAR measurements are not available.

Analysis of Depth Pro’s output shows that while depth values in the sky region may vary significantly (e.g. between 2,224m and 10,000m as shown in Figure~\ref{fig3}b), the exact metric value at these extreme distances is not critical for fog simulation as long as the region remains beyond the visibility threshold. According to the Federal Meteorological Handbook, fog is defined by a decrease in visibility to less than 1 km~\cite{r50}. Therefore, in any severity of fog, any region exceeding the 1,000m threshold will be completely occluded. So, in the case of the difference between the 2,224m sky pixel and the 10,000m sky pixel, both pixels would have the same density of fog. This makes monocular depth estimation a more suitable choice for fog simulation as it allows distant regions to be treated as beyond the effective visibility range.

\subsubsection{Quantitative Comparison}
To evaluate the trade-offs between depth completion and monocular depth estimation, a quantitative analysis was conducted on the validation set of the KITTI depth prediction dataset~\cite{r51}. The experiment compared the selected Depth Pro model against Marigold-DC using an NVIDIA Tesla V100 GPU with full FP32 precision. Predictions were clipped at a maximum distance of 120m to align with the effective range of the KITTI LiDAR sensor. The results, as summarised in~\ref{tab1}, highlight the trade-off between local depth accuracy and the suitability of each method for large-scale fog simulation.

\begin{table}[t]
\centering
\caption{Quantitative comparison of depth completion (Marigold-DC \cite{r48}) and metric depth estimation (Depth Pro \cite{r49}) on the validation set of the KITTI depth prediction dataset \cite{r51}. Predictions are clipped at a maximum distance of 120m, and the best results are made bold.}
\small
\setlength{\tabcolsep}{4pt}
\begin{tabularx}{\textwidth}{lccccccc}
\toprule
Model & \begin{tabular}[c]{@{}c@{}}RMSE\\ (m) $\downarrow$\end{tabular} & Rel $\downarrow$ & $\delta_1 \uparrow$ & $\delta_2 \uparrow$ & $\delta_3 \uparrow$ & \begin{tabular}[c]{@{}c@{}}Average\\ Inference\\ Time\\ (s)\end{tabular} & \begin{tabular}[c]{@{}c@{}}Peak\\ VRAM\\ (GB)\end{tabular} \\ \midrule
Marigold-DC \cite{r48} & \textbf{1.8918} & \textbf{0.0530} & \textbf{0.9714} & \textbf{0.9903} & 0.9960 & 25.69 & $\sim$16 \\ \addlinespace[0.5em]
Depth Pro \cite{r49}   & 3.9232 & 0.1537 & 0.8079 & 0.9836 & \textbf{0.9968} & 1.1 & $\sim$12 \\ \bottomrule
\end{tabularx}
\label{tab1}
\end{table}

As expected, Marigold-DC achieves superior local depth accuracy within the evaluated range. Its Root Mean Square Error (RMSE) is less than half that of Depth Pro, and its relative error is similarly lower. This is further reflected in the $\delta1$ accuracy metric where 97\% of Marigold-DC’s predicted pixels fall within a 25\% error margin of the ground truth compared to approximately 81\% for Depth Pro. This superior local accuracy stems from Marigold-DC leveraging the ground-truth sparse depth map as an optimisation guide.

However, these metrics fail to capture a key limitation of depth completion methods, which lies in how they handle pixels beyond the ground truth range. The inaccurate depth estimation in these out-of-range regions represents a limitation for realistic fog simulation that local metric accuracy cannot compensate for. Furthermore, Depth Pro provides substantially improved computational efficiency as it is approximately 23 times faster during inference and consumes 25\% less VRAM than Marigold-DC. The combination of this efficiency and the qualitative advantage in producing semantically consistent depth maps in distant regions makes Depth Pro a more suitable choice for the C2F pipeline.

\subsection{Camera Fog Simulation}

\subsubsection{Theoretical Model}
The C2F pipeline utilises the standard optical model based on Koschmeider’s Law~\cite{r24} as applied by Sakaridis et al.~\cite{r16}. To obtain a synthesised foggy image $I$ at pixel $x$, the model follows:
\begin{equation}
I(x)=J(x) t(x)+A(1-t(x))\,,
\label{eq1}
\end{equation}

where $J$ is the clear input image, $A$ is the atmospheric light that denotes the ambient glow added by the fog particles when light scatters off them and $t$ is the transmission map that represents the percentage of light from the clear image that passes through the fog to reach the camera at pixel $x$. Assuming a homogenous fog, the transmission $t$ is modelled as an exponential function of the scene depth $d(x)$ and an attenuation coefficient $\beta$:
\begin{equation}
t(x)=\exp ^{-\beta d(x)}\,.
\label{eq2}
\end{equation}

The coefficient $\beta$ controls the density of fog where a larger value represents thicker fog. In meteorological terms, the Meteorological Optical Range (MOR), which is also known as the visibility, is defined as the distance where the transmission $t$ is $\geq 0.05$~\cite{r52}. Using Equation~\ref{eq2}, this implies that:
\begin{equation}
\operatorname{MOR}(\text { visibility })=-\frac{\ln (0.05)}{\beta} \approx \frac{3}{\beta}\,.
\label{eq3}
\end{equation}

As mentioned earlier, fog is officially defined by a decrease in visibility to less than 1 km~\cite{r50}. Therefore, the minimum value the attenuation coefficient $\beta$ can be set as is $3 \times 10^{-3} \mathrm{~m}^{-1}$.

\subsubsection{Atmospheric Light Estimation}
\begin{figure}[t]
    \centering
    \includegraphics[width=\textwidth]{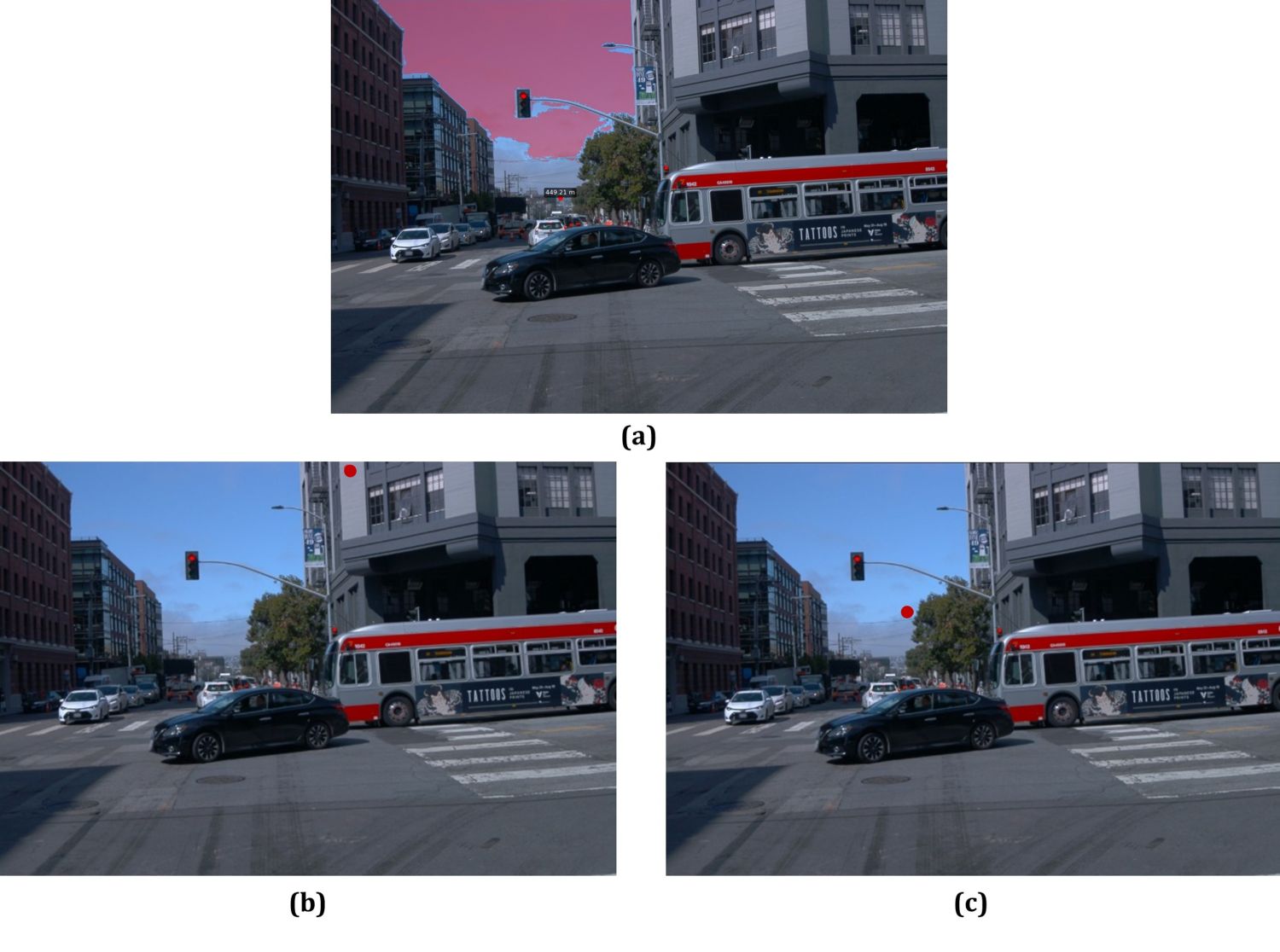}
    \caption{Depth-based atmospheric light estimation on a sample frame from the Waymo Open Dataset~\cite{r14}. (a) Identification of candidate pixels ($d > 1000m$) shown in red, which isolate the sky region. (b) Sampling without depth filtering, resulting in incorrect atmospheric light selection from a foreground building (red circle). (c) Sampling with the proposed depth-based mask, which successfully selects a representative sky pixel for accurate ambient light estimation (red circle).}
    \label{fig4}
\end{figure}

The atmospheric light A is a key parameter in the optical model as it determines the brightness and colour of the simulated fog. We initially sample A by modifying a dark channel prior method~\cite{r53} to consider only pixels beyond a 1,000m depth threshold. This ensures that the sampled atmospheric light value reflects the sky’s ambient illumination rather than midground or foreground objects as demonstrated in Figure~\ref{fig4}.

The 1,000m threshold is mathematically anchored to the Meteorological Optical Range (MOR) introduced in Equation~\ref{eq3}, which represents the boundary where clear atmospheric transmission terminates under foggy constraints. Setting the depth mask at this meteorological boundary serves as a buffer that ensures foreground and midground objects, such as vehicles, nearby buildings and roadside infrastructure, are excluded from ambient light sampling. While structures like mountains and skyscrapers may exist on the distant horizon beyond 1,000m, objects past this threshold naturally experience atmospheric haze and light scattering, causing their pixel values to mirror the true ambient light of the sky. Therefore, pixels sampled beyond this threshold provide a more reliable representation of the scene’s ambient illumination for atmospheric light estimation.

While depth-filtering improves the structural accuracy of atmospheric light estimation, sampling directly from clear sky regions often introduce an unnatural blue colour cast. This is physically inconsistent with the visual properties of fog as real-world fog is composed of relatively large water droplets that cause Mie scattering. Unlike the Rayleigh scattering that makes the clear sky appear blue, Mie scattering is not strongly wavelength-dependent, which means that it scatters all colours of visible light roughly equally with the perceptual result being that fog and clouds appear neutral in colour~\cite{r54}.

To address this chromatic bias, the C2F pipeline utilises a hybrid approach grounded in real-world data. We conducted an analysis of over 2,000 daytime foggy images from the STF~\cite{r6} and ACDC~\cite{r55} datasets to establish a realistic target range for atmospheric light. The average values are presented in Table~\ref{tab2}.

\begin{table}[t]
\centering
\caption{Empirically derived average atmospheric light values from real-world datasets.}
\begin{tabular}{lccc}
\toprule
Dataset & Average R & Average G & Average B \\ \midrule
STF \cite{r6}    & 0.6370 & 0.6374 & 0.6384 \\
ACDC \cite{r55}  & 0.8537 & 0.8531 & 0.8848 \\ \bottomrule
\end{tabular}
\label{tab2}
\end{table}

This analysis also supports the principle of Mie Scattering as the R, G and B values are nearly identical, confirming that real-world fog is spectrally neutral. To derive a realistic atmospheric light value $A$, we establish a target luminance range for daytime fog as [0.6374, 0.8555]. This range is derived by applying the colour channel averages from Table~\ref{tab2} for STF (lower bound) and ACDC (upper bound) and applying them to the following ITU-R BT.709 relative luminance formula:
\begin{equation}
\text { Luminance }=0.2126(R)+0.7152(G)+0.0722(B)\,.
\end{equation}

The final implementation for estimating the atmospheric light value A for a given image can be summarised into a three-step process:
\begin{enumerate}
    \item An initial atmospheric light value is estimated using the proposed depth-filtered dark channel prior method.
    \item The luminance of this estimated value is calculated and then clipped to the target range of [0.6374, 0.8555].
    \item A final atmospheric light vector is constructed by applying this clipped luminance value to all three colour channels.
\end{enumerate}

\begin{figure}[t]
    \centering
    \includegraphics[width=\textwidth]{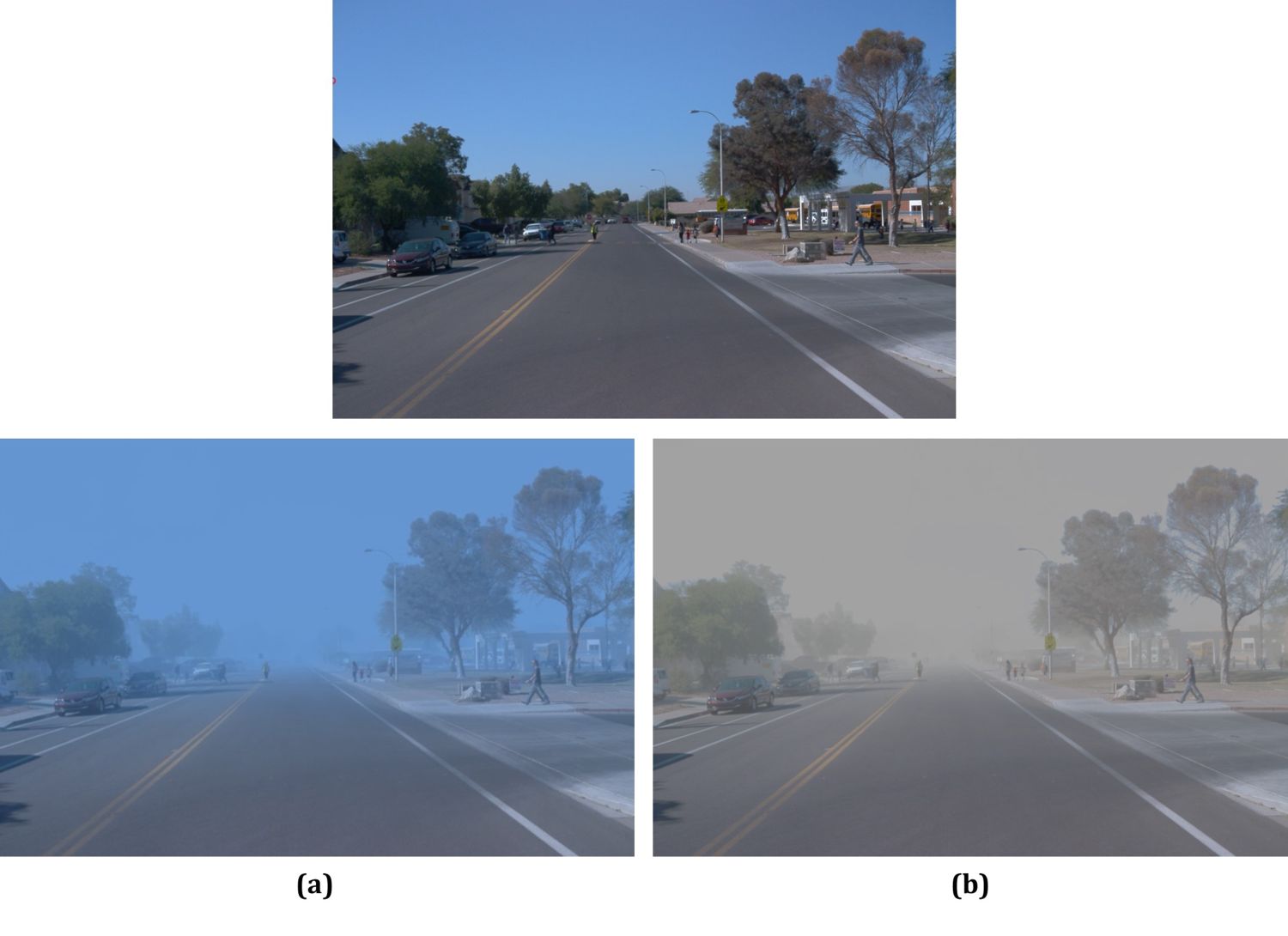}
    \caption{Visual effect of luminance-clipping method on a frame (top) from the Waymo Open Dataset~\cite{r14}. (a) Fog simulation using the depth-filtered dark channel prior method only. (b) Fog simulation using the luminance-clipping method. Fog visibility is set to 100m.}
    \label{fig5}
\end{figure}

As seen in Figure~\ref{fig5}, this method produces a more neutral fog colour that is consistent with Mie scattering principles and substantially reduces the unnatural, blue-tinted appearance that comes with sampling directly from clear sky regions.

\subsection{LiDAR Fog Simulation}
To simulate fog on the LiDAR point clouds, the C2F pipeline adopts the framework proposed by Hahner et al.~\cite{r38}, which is based on~\cite{r37}. It models attenuation and scattering effects on the LiDAR pulses.

The model simulates the power of the LiDAR return signal $P_{R}(R)$ received from a distance $R$. This received power is a combination of the power reflected from a target object $P_{R,fog}^{hard}(R)$ and the power backscattered by the fog particles $P_{R,fog}^{soft}(R)$, which manifests as noise. The equation used to model this is:
\begin{equation}
P_R(R)=P_{R, \text { fog }}^{\text {hard }}(R)+P_{R, \text { fog }}^{\text {soft }}(R)\,.
\end{equation}

The attenuation effect is captured in the $P_{R,fog}^{hard}(R)$ term. As the laser pulse travels from the sensor to an object, its energy is reduced by the fog, and the reflected pulse is attenuated again on its return journey. This two-way attenuation means that in foggy conditions, objects appear to have a lower intensity, and distant objects may completely disappear if their returned signal is too weak to be detected.

The backscattering effect $P_{R,fog}^{soft}(R)$ models the soft target return. It describes the phenomenon where the laser pulse is reflected back to the sensor by the fog particles suspended in the air. This creates a noisy veil of ghost points, which is noticeable at close ranges and can obscure real objects. The model calculates this term by integrating the effect of all fog particles along the laser's path.

The final implementation is determined by the maximum power return. If the backscattered noise is more powerful than the attenuated hard target for a given point, it is relocated to a closer range corresponding to the fog’s peak reflection, effectively generating a phantom point. However, if the attenuated hard target remains stronger, the point’s original spatial location is preserved, but its intensity is reduced to reflect the energy lost during transmission.

The algorithm developed by Hahner et al.~\cite{r38} provides a simple method for simulating fog on clear-weather LiDAR point clouds. As input, the algorithm requires a clear point cloud $(x,y,z)$, the measured intensity $i$ of the point, the extinction coefficient $\alpha$, the backscattering coefficient $\beta$, the differential reflectivity of the surface $\beta_{0}$ and the half-power pulse width $\tau_{H}$. Table~\ref{tab3} summarises the inputs of the algorithm.

\begin{table}[t]
\centering
\caption{The LiDAR sensor parameters used to configure the fog simulation model, based on \cite{r37}, \cite{r38}.}
\renewcommand{\arraystretch}{2} 
\begin{tabular}{lc}
\toprule
\multicolumn{1}{c}{Input} & \multicolumn{1}{c}{Formula} \\ \midrule
Clear Point Cloud & $(x, y, z)$ \\
Measured Intensity of the Point & $i$ \\
Extinction Coefficient & $\alpha = \dfrac{3}{MOR}$ \\
Backscattering Coefficient & $\beta = \dfrac{0.046}{MOR}$ \\
Differential Reflectivity & $\beta_0 = \dfrac{1 \times 10^{-6}}{\pi}$ \\
Half-Power Pulse Width & $\tau_H = 20 \, ns$ \\ \bottomrule
\end{tabular}
\label{tab3}
\end{table}

\subsection{Qualitative Validation of the Clear2Fog Pipeline}

\subsubsection{Performance on the Waymo Open Dataset}
\begin{figure}[t]
    \centering
    \includegraphics[width=\textwidth]{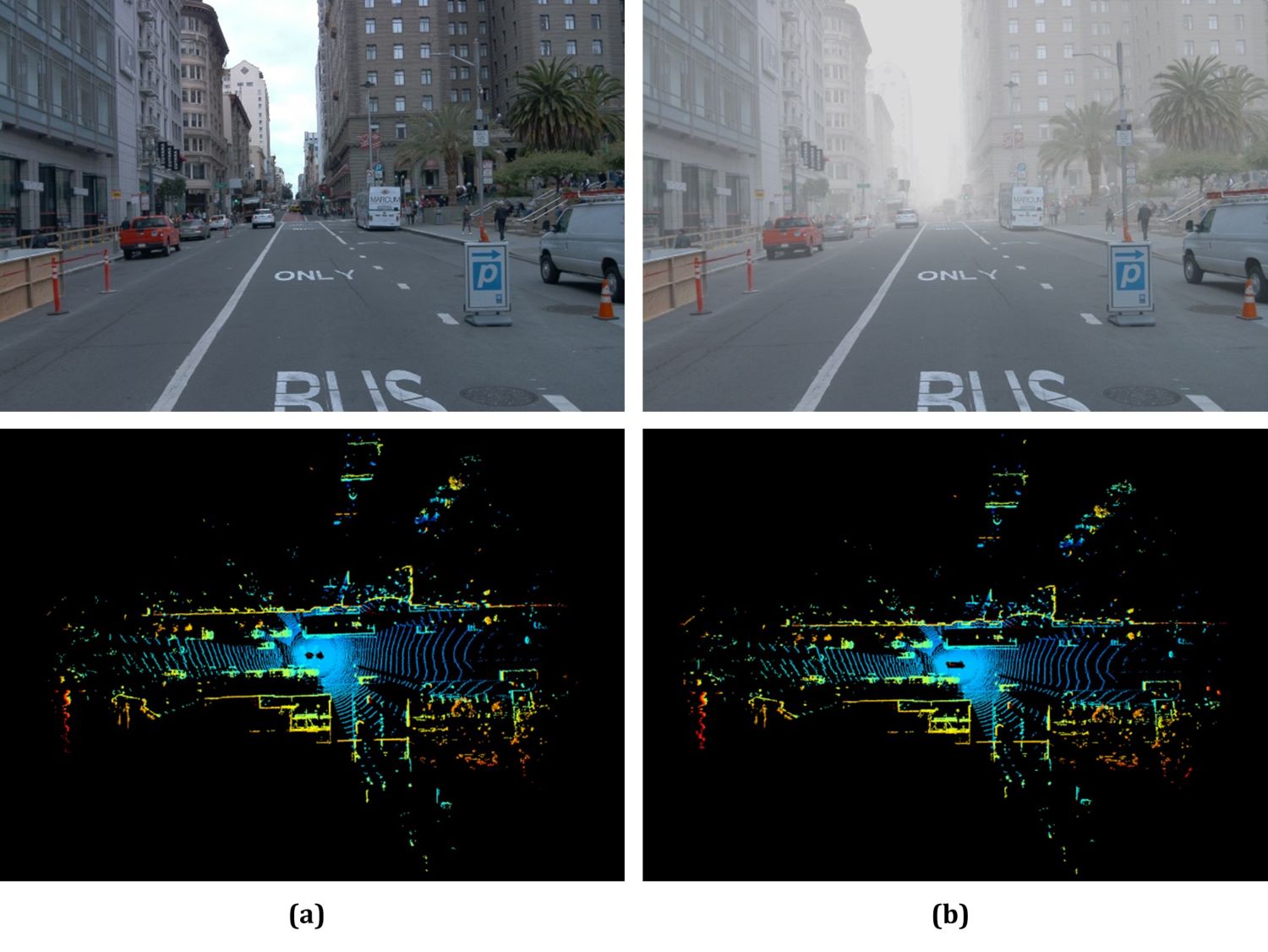}
    \caption{Validating the C2F pipeline on the Waymo Open Dataset~\cite{r14}. (a) Original clear-weather data with a camera view (top) and its corresponding LiDAR point cloud (bottom). (b) The foggy output generated by the pipeline using a fog visibility parameter of 150m.}
    \label{fig6}
\end{figure}

We first validate the pipeline’s end-to-end functionality using a representative frame from the Waymo Open Dataset~\cite{r14} to assess the synchronisation between modalities. Figure~\ref{fig6} presents the front camera view for a sample frame alongside its corresponding LiDAR point cloud, comparing the original clear-weather data with the foggy output generated by the C2F pipeline. 

The results illustrate the pipeline’s ability to generate consistent and physically plausible fog across both sensor modalities. In the camera image, this is demonstrated through a progressive reduction in visibility, particularly in the distant regions of the scene, which is consistent with the underlying atmospheric scattering model. In the LiDAR point cloud, the impact of fog differs because active LiDAR sensors interact with fog through laser backscatter and signal attenuation rather than visual reduction. Consequently, nearby scene structure is largely preserved, while distant areas experience signal attenuation and backscatter with the addition of phantom points. Together, these results indicate that the pipeline captures the distinct effects of fog on camera and LiDAR sensors.

\subsubsection{Generalisation to 2D Image Datasets}
\begin{figure}
    \centering
    \includegraphics[width=\textwidth]{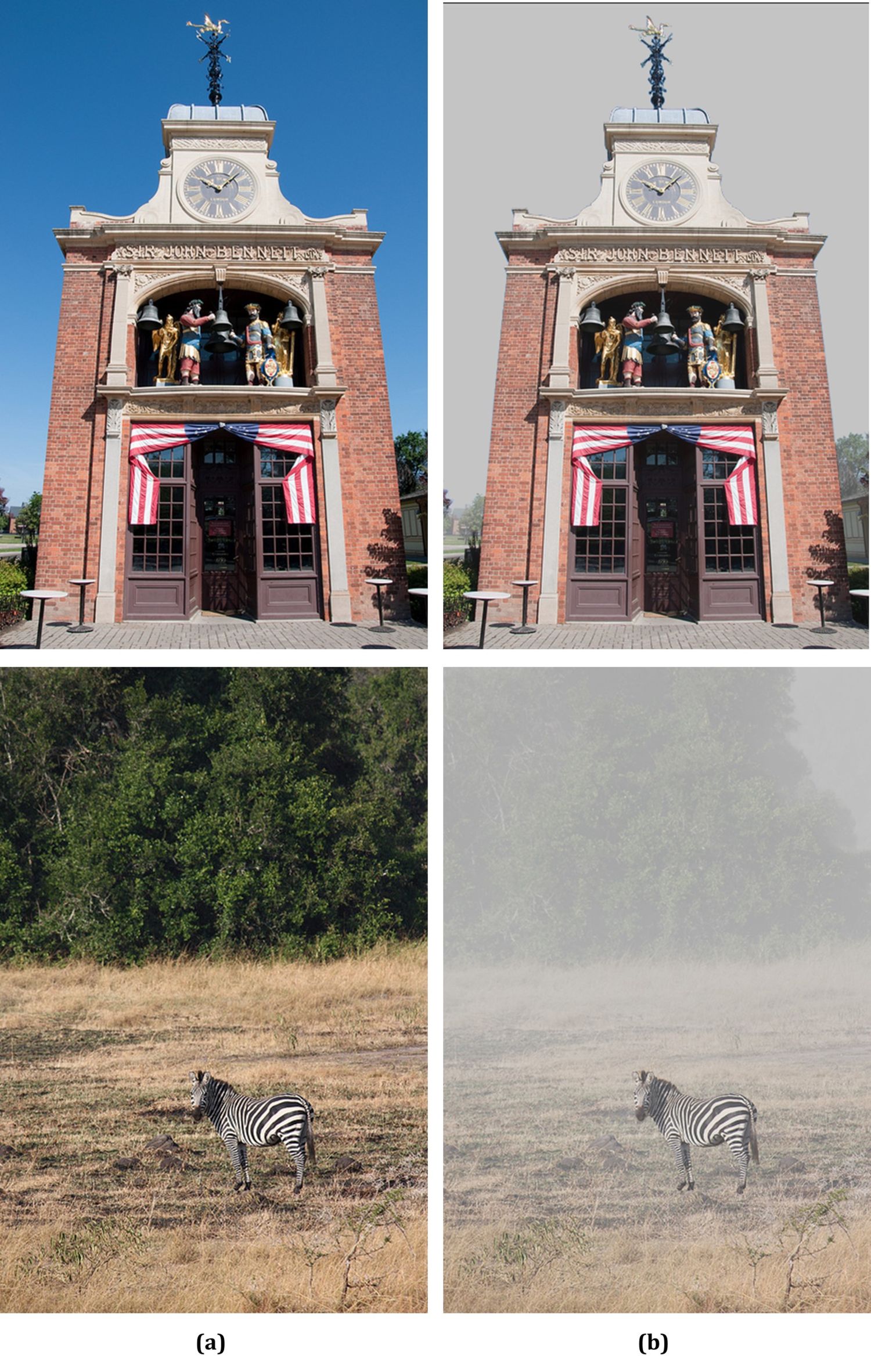}
    \caption{C2F application on the COCO 2017 dataset~\cite{r56}. (a) Displays the original clear-weather images. (b) Displays the foggy output from the pipeline, which was generated using a fog visibility parameter of 150m.}
    \label{fig7}
\end{figure}

\begin{figure}
    \centering
    \includegraphics[width=\textwidth]{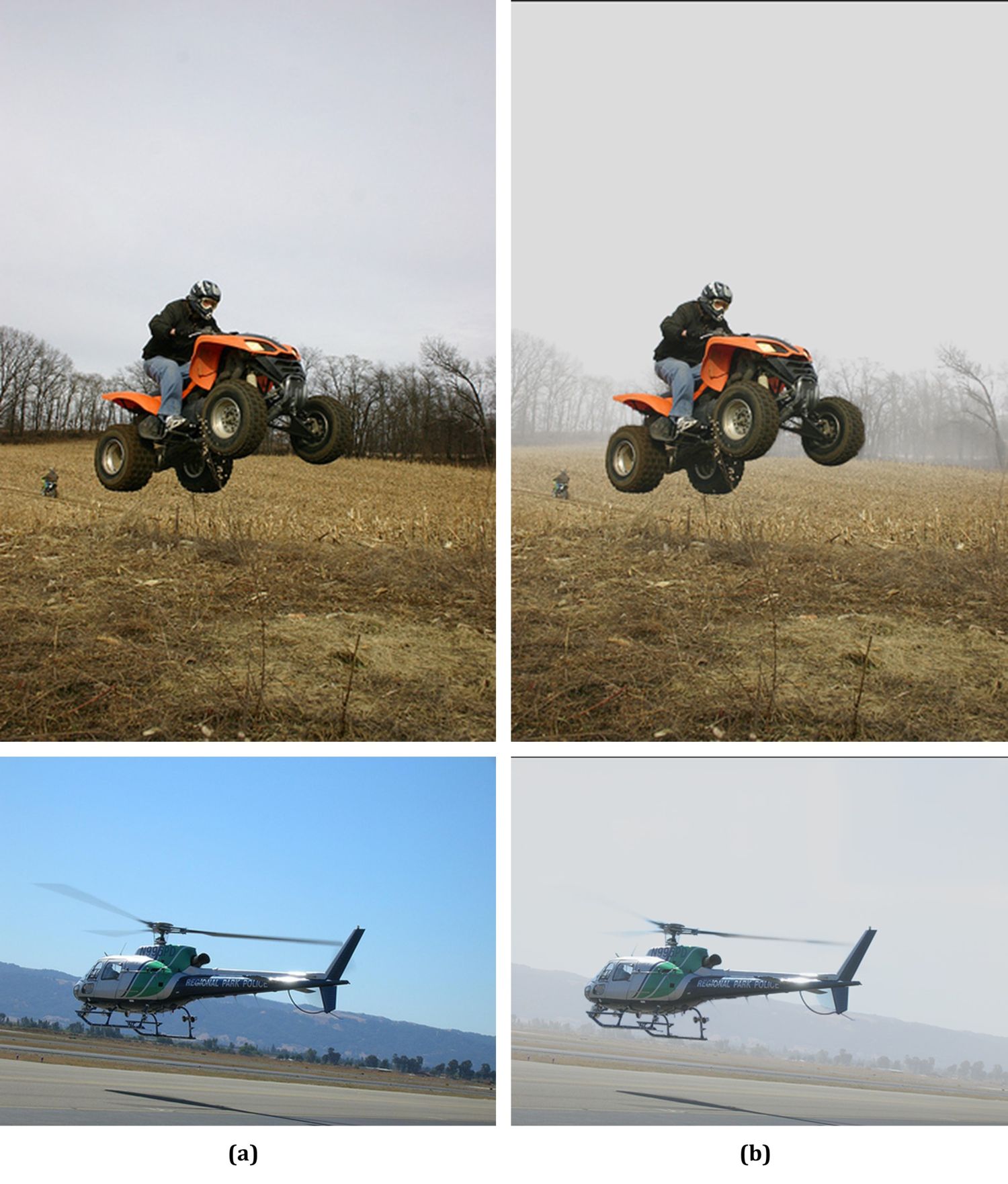}
    \caption{C2F application on the Flickr30k dataset~\cite{r57}. (a) Displays the original clear-weather images. (b) Displays the foggy output from the pipeline, which was generated using a fog visibility parameter of 150m.}
    \label{fig8}
\end{figure}

To demonstrate the generalisability of the C2F pipeline beyond the autonomous driving domain, we apply the framework to samples from standard 2D image datasets, specifically COCO 2017~\cite{r56} and Flickr30k~\cite{r57}. This evaluation tests the pipeline’s ability to handle diverse resolutions, scene compositions and lighting conditions that contrast with the more structured nature of autonomous driving datasets. As shown in Figure~\ref{fig7} and Figure~\ref{fig8}, the pipeline generates realistic fog across these varied conditions. This cross-domain application is made possible with the integration of a monocular metric depth estimation model, which allows for a more realistic fog simulation, even in the absence of LiDAR data. The results show that the design of the pipeline is not overfitted to a specific data source or context, creating a flexible tool that is not restricted to autonomous driving datasets.

\bigskip
\section{Experiments}
\label{sec:experiments}
The primary objective of this section is to investigate how dataset scale and environmental diversity in synthetic fog data influence object detection performance. While the Clear2Fog (C2F) pipeline is used for data generation, this evaluation focuses on whether increasing the size of synthetic data or expanding its diversity through varying fog densities provides greater benefits under synthetic and real-world fog conditions. Through this systematic data efficiency study, we assess the effectiveness of large-scale synthetic datasets for adverse weather perception and analyse the factors that contribute to improved model generalisation.

\subsection{Experimental Settings}
The study utilises a subset of the Waymo Open Dataset (v1.4.3)~\cite{r14}. To provide a consistent clear-weather baseline for fog generation, we filtered the dataset to include only daytime scenes captured in clear weather conditions. Specifically, we used a training set consisting of a randomly sampled subset of 270 scenes ($\sim$270,000 images) from the official training split and reserved a separate hold-out set of 30 scenes ($\sim$30,000 images) for validation. To ensure access to ground-truth annotations, we randomly selected a test set of 150 scenes ($\sim$150,000 images) from the official validation set. For fog simulation, we used the C2F pipeline to generate two training sets:
\begin{enumerate}
    \item Fixed-density dataset: A uniform distribution where we generated all images with a fog visibility of 150m, representing moderate fog conditions.
    \item Mixed-density dataset: A stratified distribution where we assigned each scene a visibility from one of five levels: 50m, 100m, 150m, 200m and 300m. This distribution reflects real-world constraints where visibility below 50m represents challenging visibility conditions, whereas visibility above 300m presents only mild haze that standard camera sensors can manage. The intermediate steps create a balanced progression of fog severity, reducing the risk of over-specialising to a single fog level.
\end{enumerate}

To quantify the marginal gains of data scale, we trained the models on five distinct subsets: 10\%, 25\%, 50\%, 75\% and 100\% of the available training data. All foggy images inherited the original, unchanged 2D bounding box annotations from the original dataset.

The primary model used for our experiments was a Faster R-CNN [60] built on a ResNet-50 backbone from the MMDetection library~\cite{r59}. To investigate whether the observed data efficiency trend extended beyond the primary architecture, we evaluated a YOLOX-S model~\cite{r60} from the same library across the same dataset scale settings. Both architectures were initialised from COCO pre-trained weights. To evaluate performance consistency across training runs, each experiment was repeated using three different random seeds (12, 34 and 56). The reported metrics represent the average Mean Average Precision (mAP) and standard deviation across the three primary classes: Vehicles, Pedestrians and Cyclists.

For experiments involving fine-tuning synthetic models on real-world foggy data, we used the best-performing epoch from each synthetic training seed to provide a consistent starting point for domain transfer. To isolate the effects of the learning rate strategy on sim-to-real performance, we conducted these subsequent fine-tuning stages using a single representative seed.

To evaluate sim-to-real transfer under real fog conditions, we constructed a fog subset from the Seeing Through Fog (STF) dataset~\cite{r6}. We utilised a total of 1,140 left-camera RGB images containing 3,362 annotations to form the STF-Foggy dataset and mapped the annotations to follow the same taxonomy used throughout this work (Vehicles, Pedestrians and Cyclists), allowing direct comparison with the models trained on the synthetic data. We split the STF-Foggy dataset by scene into training, validation and test sets using an approximate 70/15/15 ratio; this ensured that frames from the same scene did not appear across multiple splits. Although STF originates from a different geographical domain than Waymo, it provides a challenging cross-domain evaluation setting where all models are evaluated under identical real-world conditions. The relatively small size of STF-Foggy directly highlights the severe scarcity of annotated, real-world foggy data available to researchers. While massive scale is easily achievable in synthetic environments, real-world validation is fundamentally limited by what’s publicly available, which motivates the need for scalable synthetic data generation approaches. We performed all experiments on the NVIDIA L40S GPU and have provided a list of all the specific scenes used throughout this study on the GitHub repository.

\subsection{Perceptual Realism and Atmospheric Light Analysis}
To evaluate the physical and functional accuracy of the Clear2Fog (C2F) pipeline, we conducted a combined evaluation that compares human perceptual preference with quantitative object detection performance. The human study evaluates the impact of the complete simulation pipeline, including differences in depth representation and atmospheric light estimation, under matched fog visibility conditions.

\subsubsection{Human Perceptual Study}
Synthetic fog primarily alters global image appearance by fading object details, shifting colour balances and obscuring objects relative to their distance from the camera. While object detection performance can indicate the usefulness of a synthetic dataset for model training, it cannot measure perceptual realism. There is currently a lack of a widely accepted quantitative metric to evaluate fog realism for autonomous driving data. While reference-less frameworks based on Natural Scene Statistics (NSS) like AuthESI~\cite{r29} exist, they are optimised for generic photography and natural landscapes as opposed to the highly structured layouts of driving scenes. Because we cannot rely fully on such metrics to capture how humans perceive environmental realism in the autonomous driving domain, we use a human perceptual study to provide a supporting qualitative indicator of our pipeline's visual realism.

We evaluated visual realism by comparing the images generated by our C2F pipeline against those provided in the Multifog KITTI dataset~\cite{r7}. The Multifog KITTI dataset consists of 7,481 foggy images with fog densities ranging from 20m to 80m. It was selected as a representative baseline that adapts the established Foggy Cityscapes~\cite{r16} methodology to a multimodal format and utilises a more contemporary depth completion method. Since many physics-based approaches share the same underlying Koschmieder optical model~\cite{r24}, we selected Multifog KITTI as a representative baseline. Furthermore, we intentionally exclude generative learning models from this study as they do not generally provide the same explicit control of the visibility parameters, whereas the study requires measurable control over this setting.

For the human perceptual study, we randomly sampled 20 images from the clear KITTI dataset~\cite{r17} and presented them in a blind forced-choice setup to 22 non-expert participants to rely on standard human visual intuition rather than specialised computer vision expertise. To ensure robustness across different scene types, all 20 images were randomly sampled to represent independent scenes, and image ordering was randomised to avoid bias. For each clear image, participants were presented with a side-by-side pair of synthetic foggy images: one was generated using C2F and the other was sourced from Multifog KITTI. To remove fog density as a confounding variable, the visibility parameter for each C2F-generated image was mapped directly on a frame-by-frame basis to match the exact corresponding density level (from 20m to 80m) featured in the Multifog KITTI version. The participants were asked to identify which image showed a more realistic simulation of fog based on the clear image, and across a total trial volume of 440 total pairwise judgements, the C2F-generated images were selected as more realistic 92.95\% of the time. A two-sided binomial test against a random-choice baseline ($p=0.5$) confirms that this preference is statistically significant ($p<0.0001$). To measure rater reliability across this evaluation scale, we calculated the inter-rater agreement across all judges utilising a free-marginal multirater Kappa ($\kappa_{\text{free}}$), yielding a value of $\kappa_{\text{free}} = 0.75$. This indicates a statistically substantial level of agreement among the participants.

\begin{figure}[t]
    \centering
    \includegraphics[width=\textwidth]{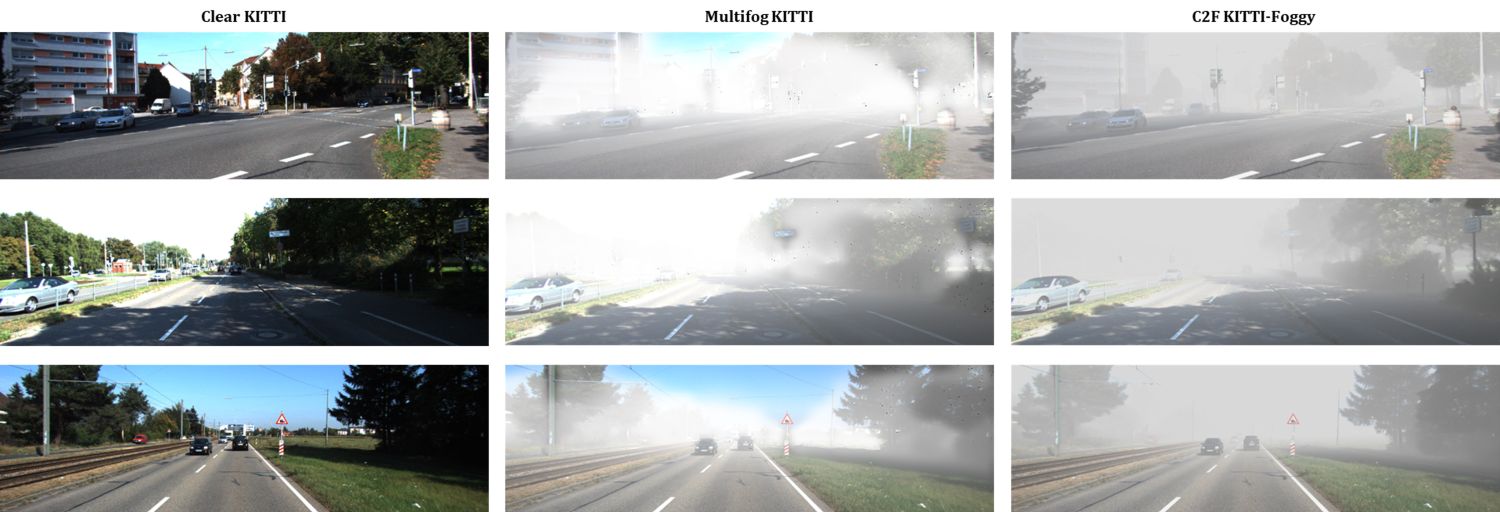}
    \caption{Qualitative comparison of fog simulation realism from the human perceptual study between Multifog KITTI~\cite{r7} and Clear2Fog (C2F).}
    \label{fig9}
\end{figure}

Figure~\ref{fig9} presents some examples from the human study. The C2F pipeline produces fog distributions that were preferred by participants, especially in the distant regions and sky areas. Also, the atmospheric light appears more diffused and neutral, avoiding the over-brightening present in the baseline model. Since both frameworks are based on the same underlying optical scattering model, the observed perceptual differences are mainly due to the combined effects of depth and atmospheric light estimations. So, to detach perceptual realism from detection performance, we present a controlled detection-based analysis next.

\subsubsection{Quantitative Evaluation and Atmospheric Light Ablation}
To examine whether perceptual realism translates into improved downstream performance, we conducted a controlled quantitative study using object detection. The C2F pipeline was compared to Multifog KITTI under controlled conditions where both frameworks used the same depth estimation model (Depth Pro) to evaluate the impact of differences in atmospheric light estimation. We created two versions of the KITTI dataset, named KITTI-Foggy, using the C2F pipeline; one version was created with a fixed-density fog of 50m and the other was created with mixed-density fog between 20m-80m to match the parameters of the Multifog KITTI dataset. We fine-tuned all models on the training set of STF-Foggy and tested it on its test set.

\begin{table}[t]
\centering
\caption{Atmospheric light ablation study under matched depth estimation conditions.}
\begin{tabularx}{\textwidth}{lX}
\toprule
Pre-Training Condition & \multicolumn{1}{c}{Average mAP $\pm$ Std. Dev.} \\ \midrule
None (Real-Only Baseline) & \multicolumn{1}{c}{$0.2553 \pm 0.0023$} \\ \addlinespace[0.3em]
KITTI-Foggy Fixed-Density & \multicolumn{1}{c}{$0.2587 \pm 0.0107$} \\ \addlinespace[0.3em]
KITTI-Foggy Mixed-Density & \multicolumn{1}{c}{$0.2720 \pm 0.0139$} \\ \addlinespace[0.3em]
Multifog KITTI \cite{r7} (using Depth Pro) & \multicolumn{1}{c}{0.2787 $\pm$ 0.0107} \\ \bottomrule
\end{tabularx}
\label{tab4}
\end{table}

The results from Table~\ref{tab4} show that synthetic pre-training improves mean detection performance compared with the real-only baseline. However, due to the variance observed across the independent seeds, the performance distributions of the three pre-training conditions exhibit statistical overlap. While Multifog KITTI achieves the highest absolute mean score, its variance overlaps with that of the KITTI-Foggy mixed-density variant.  Therefore, no statistically significant performance difference can be established between these two approaches based solely on mAP. This overlap indicates that both atmospheric light frameworks result in comparable downstream detection performance, indicating that the increased perceptual realism of C2F does not result in degraded detection performance.

\subsubsection{Discussion}
Taking the perceptual and quantitative results together, these findings suggest that perceptual realism and downstream task performance are not necessarily aligned. Although the proposed C2F pipeline is overwhelmingly preferred by humans due to its perceived realism, this perceptual advantage does not translate into a statistically significant improvement in object detection performance under the evaluated setting. This discrepancy likely arises as human perception and object detectors rely on different visual cues. While humans judge realism based on global appearance statistics, such as colour and scene consistency, detectors mainly learn task-specific features relevant for object recognition. Therefore, a simulation framework that introduces unnatural chromatic biases can provide the same detection cues as a physically realistic framework, resulting in statistically comparable mAP outputs despite differences in visual appearance.

This observation has implications related to the field of synthetic fog generation. It demonstrates that major advancements in visual and structural realism can be achieved without sacrificing downstream performance. While detection performance is the main objective for synthetic training, improving perceptual realism is important in reducing artefacts and improving the reliability of the data, especially in safety-critical applications. By eliminating chromatic and structural artefacts, the C2F pipeline ensures that the generated data is highly interpretable and visually plausible in scenarios where human decisions are important considerations alongside raw quantitative performance metrics.

\subsection{Data Efficiency and Environmental Diversity Study}
This section analyses the impact of dataset size and environmental diversity on object detection performance within synthetic fog environments. To comprehensively assess these two factors, we tested our models across three validation sets created from the 150-scene Waymo test split: a clear-weather baseline, a fixed-density fog set (150m visibility) and a mixed-density fog set (50m-300m visibility). For each validation set, we evaluated three separate training configurations: a model trained exclusively on clear weather, a model trained on fixed-density fog and a model trained on mixed-density fog. In order to examine whether the observed trends were consistent across two representative detector architectures, each of these three training configurations was evaluated twice: once using the two-stage Faster R-CNN~\cite{r58} and the other using the one-stage YOLOX-S~\cite{r60}. All reported metrics represent the mean values and standard deviations calculated across three independent training seeds.

\begin{figure}
    \centering
    \includegraphics[width=\textwidth]{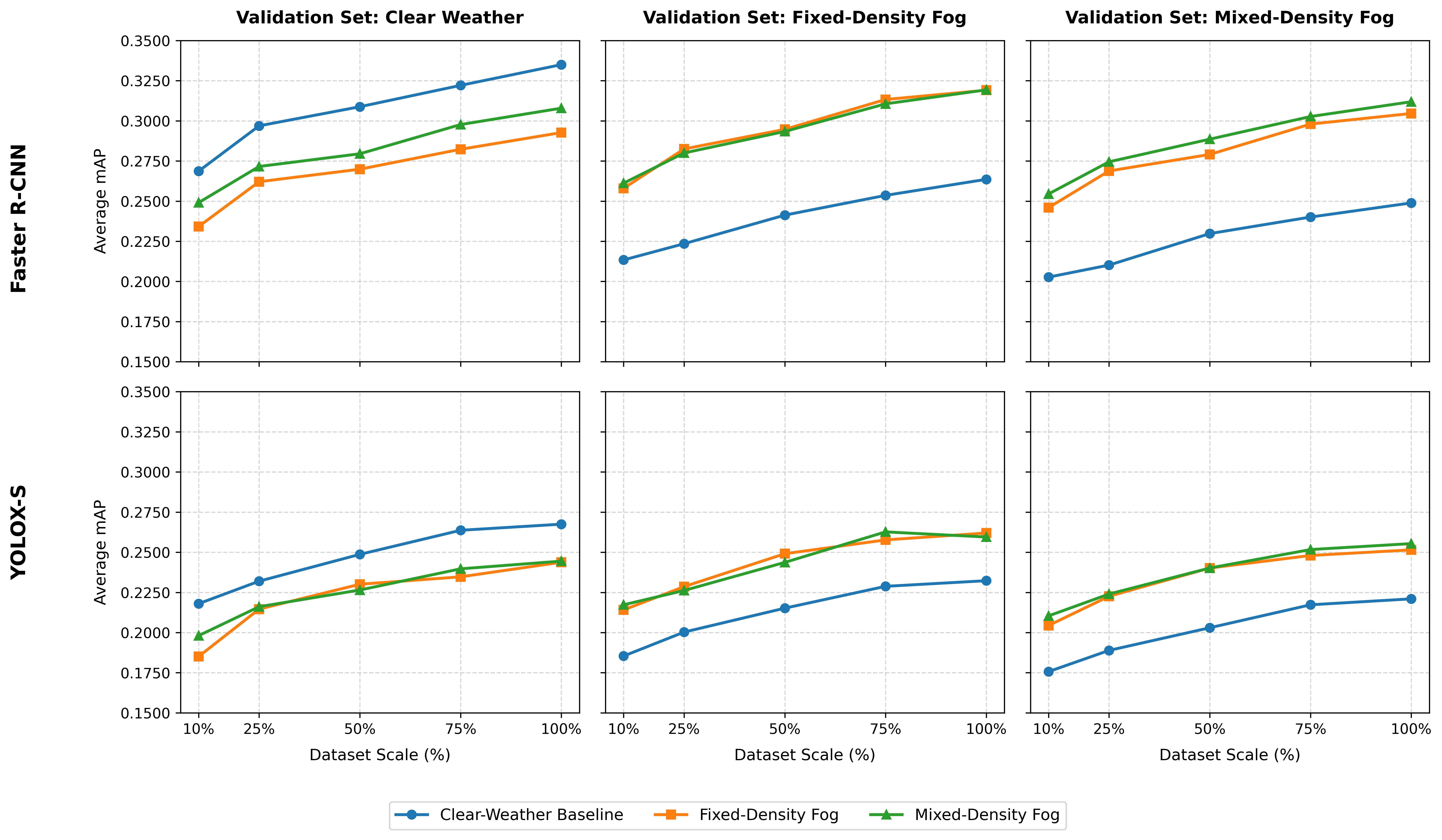}
    \caption{Object detection performance (mAP) showing scaling trends across dataset subsets (10\% to 100\%) for Faster R-CNN~\cite{r58} (top row) and YOLOX-S~\cite{r60} (bottom row). Columns represent Clear Weather (left), Fixed-Density Fog (centre) and Mixed-Density Fog (right) validation environments. Data points are averaged across three independent seeds. Full class-wise and standard deviation breakdowns are provided in Appendix A, Tables~\ref{tabA1}-~\ref{tabA6}.}
    \label{fig10}
\end{figure}

As shown in Figure~\ref{fig10}, object detection performance steadily improves as the training dataset size increases across all environments. For the Faster R-CNN architecture, expanding the training data from the initial 10\% subset to the full 100\% scale yields steady gains ranging from roughly 22\% to 25\% across all three validation domains. This trend is closely mirrored by the YOLOX-S framework where expanding the training data provides steady improvements of approximately 19\% to 30\% (as detailed in Appendix A, Tables~\ref{tabA1}-~\ref{tabA6}). This consistent behaviour across both frameworks suggests that scaling up synthetic training data can be an effective approach to improve object detection accuracy within the evaluated conditions. An analysis of the full class-wise breakdowns shows that this scaling behaviour is also apparent in the individual classes presented in Tables~\ref{tabA1}-~\ref{tabA6}. For example, looking at the class-wise performance of the Faster R-CNN model on the clear validation set (Table~\ref{tabA1}), scaling the clear-weather baseline from 10\% to 100\% raises Vehicle AP from 0.3520 to 0.4260, Pedestrian AP from 0.3033 to 0.3493 and Cyclist AP from 0.1513 to 0.2297. While the smaller classes like pedestrians and cyclists naturally result in lower absolute metrics due to factors such as smaller object size and fewer instances, their consistent performance improvements follow the scaling trend observed for vehicles across both evaluated frameworks.

A notable relationship emerges when evaluating model generalisation across different weather distributions. As expected, models trained exclusively on clear-weather data achieve the highest peak performance when tested on the clear validation set for both architectures. However, between the two synthetic models, the mixed-density fog model provides stronger performances on the clear-weather validation sets compared to the fixed-density fog model, maintaining a higher mAP score across almost all data subsets. When transitioning to the two foggy test sets, the clear-weather models show a noticeable reduction in performance. On the fixed-density validation set, the mixed-density model performs on par with the fixed-density model despite the latter being trained exclusively on fixed-visibility conditions. Specifically, the 100\% scale Faster R-CNN fixed-density model achieves an average mAP of $0.3191 \pm 0.0029$, while the 100\% mixed-density model closely matches it at $0.3193 \pm 0.0025$.

The evaluation highlights a data efficiency trend: a 75\% scale mixed-density training set  achieves comparable performance to a 100\% scale fixed-density training set within the evaluated fog environments. As visualised in Figure~\ref{fig10} and detailed across Tables~\ref{tabA1}-~\ref{tabA6}, the performance of the 75\% mixed-density model performs within a narrow margin of the 100\% fixed-density model across the different validation sets for both Faster R-CNN and YOLOX-S. To evaluate the statistical significance of this trend, we performed a Welch's $t$-test ($N = 3$, $\alpha = 0.05$). The results showed no statistically significant difference across three of the four evaluated configurations: Faster R-CNN on the mixed-density validation set ($p = 0.5286$), YOLOX-S on the fixed-density validation set ($p = 0.6924$), and YOLOX-S on the mixed-density validation set ($p = 0.2406$). A statistically significant difference was observed for Faster R-CNN on the fixed-density validation set ($p = 0.0359$); however, this difference corresponds to a small absolute mAP change of approximately $0.0085$, which can be associated with the low variance observed across the repeated runs. We note that because each configuration consists of only three independent training runs, the statistical power of these tests is limited and does not inherently prove absolute equivalence.

Overall, the results suggest that within the evaluated synthetic fog environments, mixed-density training can reduce training data scale by 25\% while maintaining comparable detection performance. The consistent trends observed across both architectures indicate that increasing the diversity of fog conditions in training may provide a more data-efficient alternative to increasing dataset scale.

\subsection{Sim-to-Real Validation}
To evaluate the direct utility of synthetic data for real-world perception, we evaluated the 100\% subsets of the synthetic models against the complete dataset of STF-Foggy. This comparison tests whether a model trained exclusively on synthetic fog can transfer to real-world fog conditions.

\begin{table}[t]
\centering
\caption{Zero-shot validation of the synthetic 100\% subset models on the real images of the STF-Foggy dataset.}
\small
\begin{tabularx}{0.7\textwidth}{X c}
\toprule
\multicolumn{1}{c}{Condition} & Average mAP $\pm$ Std. Dev. \\ \midrule
\begin{tabular}[c]{@{}l@{}}Clear-Weather Baseline (real)\end{tabular} & 0.0653 $\pm$ 0.0006 \\ \addlinespace[0.5em]
\begin{tabular}[c]{@{}l@{}}Fixed-Density Fog (synthetic)\end{tabular} & $0.0630 \pm 0.0010$ \\ \addlinespace[0.5em]
\begin{tabular}[c]{@{}l@{}}Mixed-Density Fog (synthetic)\end{tabular} & $0.0590 \pm 0.0010$ \\ \bottomrule
\end{tabularx}
\label{tab9}
\end{table}

The results summarised in Table~\ref{tab9} show that the clear-weather baseline marginally outperforms both synthetic foggy models despite having no prior exposure to fog during training. This confirms that synthetic fog alone is insufficient to fully bridge the sim-to-real domain gap. These performance discrepancies may be attributed to differences between synthetic and real fog features, including potential artefacts introduced during depth estimation and simulation. Furthermore, the mixed-density model performed slightly worse than the fixed-density model, with one possible explanation being that the increased diversity of synthetic conditions also introduces a wider range of variations that do not fully match the distribution of real fog.

To determine if this performance gap was specific to the C2F pipeline, we conducted a validation test using the Multifog KITTI dataset, which utilises different methods for depth and atmospheric light estimations. We split the dataset according to the method provided by OpenMMLab~\cite{r59}. Table~\ref{tab10} shows the result of training both the clear-weather KITTI and the Multifog KITTI on the same Faster R-CNN model and testing them on STF-Foggy.

\begin{table}[t]
\centering
\caption{Zero-shot performance of Multifog KITTI \cite{r7} on real-world STF-Foggy data.}
\small
\begin{tabularx}{0.7\textwidth}{X c}
\toprule
\multicolumn{1}{c}{Condition} & Average mAP $\pm$ Std. Dev. \\ \midrule
Clear KITTI \cite{r17} & 0.0603 $\pm$ 0.0032 \\ \addlinespace[0.5em]
Multifog KITTI \cite{r7} & $0.0597 \pm 0.0015$ \\ \bottomrule
\end{tabularx}
\label{tab10}
\end{table}

As shown in Table~\ref{tab10}, the zero-shot performance of the synthetic Multifog KITTI model is comparable to the clear KITTI baseline, with their scores overlapping within a single standard deviation. These findings demonstrate that models trained exclusively on synthetic foggy images cannot yet overcome the sim-to-real gap as they fail to convincingly outperform clear-weather baselines when faced with real-world fog. This emphasises the need to explore hybrid training strategies, such as utilising large-scale synthetic data as a pre-training tool rather than a standalone replacement for real-world data. Therefore, we next investigate whether synthetic pre-training can provide stronger initialisation for real fog adaptation.

\subsection{Fine-Tuning on Real Fog}
This section investigates the utility of the C2F pipeline as a pre-training tool using the Waymo Open Dataset. We further fine-tuned the 100\% fixed-density and mixed-density synthetic models on the training set of STF-Foggy to determine if large-scale synthetic datasets provide a stronger initialisation than training on real data alone. Initially, we performed the fine-tuning using the default hyperparameters used during initial training, including a learning rate (LR) of 0.02. However, as shown in Table~\ref{tab11}, this standard setup failed to yield any performance advantage over the baseline model trained solely on the real STF-Foggy training set with the pre-trained models instead showing a consistent downward trend in mean performance.

\begin{table}[t]
\centering
\caption{Impact of synthetic pre-training on sim-to-real transfer using standard fine-tuning ($LR=0.02$).}
\small
\begin{tabularx}{0.8\textwidth}{X c}
\toprule
\multicolumn{1}{c}{Pre-Training Condition} & Average mAP $\pm$ Std. Dev. \\ \midrule
None (Real-Only Baseline) & 0.2553 $\pm$ 0.0023 \\ \addlinespace[0.5em]
C2F Fixed-Density (Waymo) & $0.2450 \pm 0.0053$ \\ \addlinespace[0.5em]
C2F Mixed-Density (Waymo) & $0.2487 \pm 0.0046$ \\ \addlinespace[0.5em]
Multifog KITTI \cite{r7}  & $0.2357 \pm 0.0092$ \\ \bottomrule
\end{tabularx}
\label{tab11}
\end{table}

The results indicate a negative transfer effect where the distribution differences between synthetic and real fog limit the effectiveness of standard fine-tuning. A possible explanation is that the relatively low optimisation step limits the model’s ability to sufficiently adapt the synthetic feature representations to real-world fog. This trend is observed across both simulation pipelines, suggesting that standard fine-tuning settings may be insufficient for improving sim-to-real adaptation. To address this, the following section investigates learning rate sensitivity to determine whether adjusting the learning rate can improve the adaptation.

\subsection{Improving Sim-to-Real Adaptation}
\begin{figure}
    \centering
    \includegraphics[width=\textwidth]{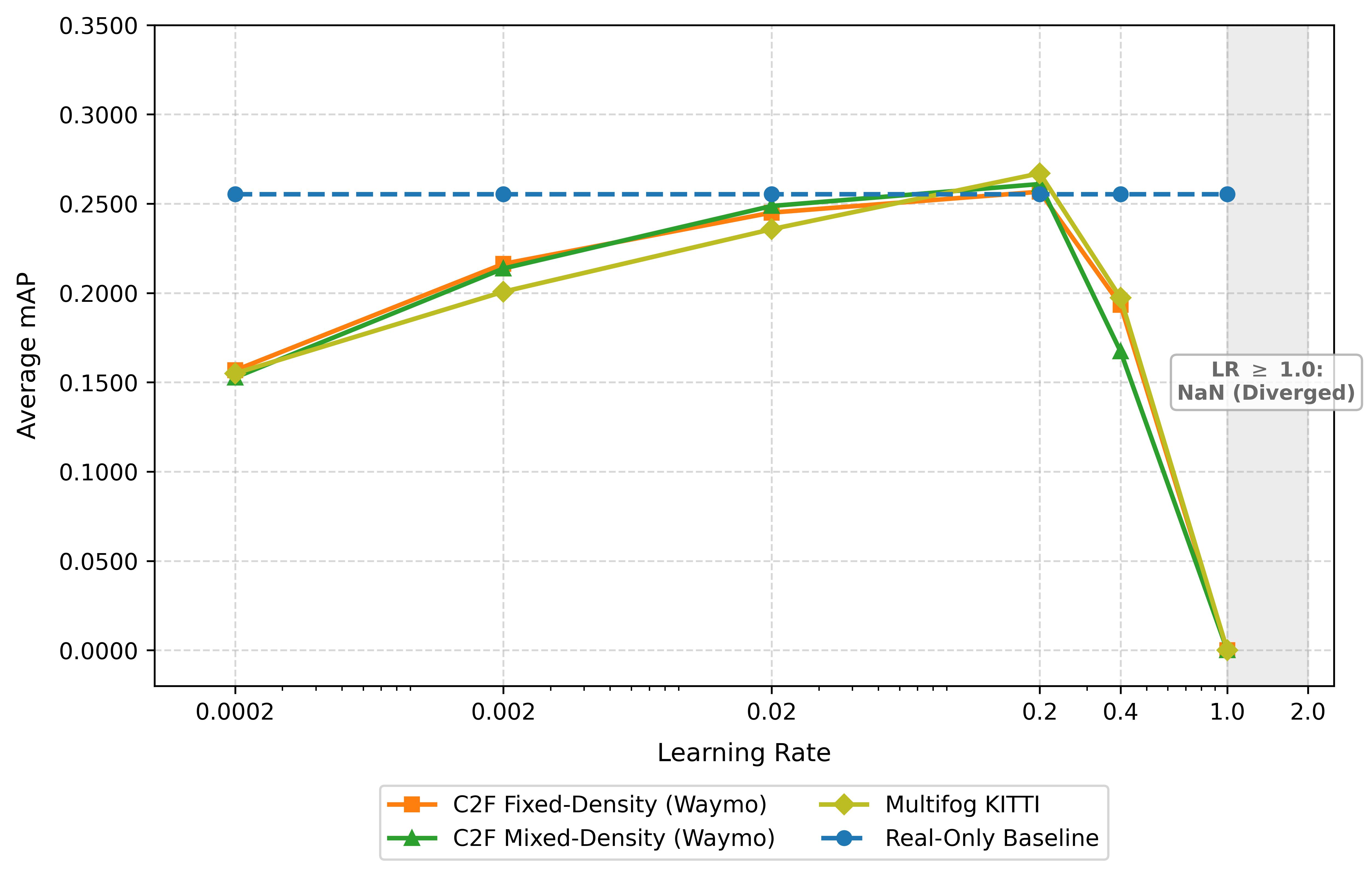}
    \caption{Learning rate sensitivity analysis on the test set of STF-Foggy. The trajectories indicate the mean mAP evaluated across three random seeds. The horizontal dashed line establishes the real-only baseline performance (0.2553 mAP). The shaded region indicates the boundary of total instability where training became unstable and produced NaN outputs.}
    \label{fig11}
\end{figure}

The initial fine-tuning experiments suggested that the default learning rate ($LR=0.02$) was insufficient for the models to effectively adapt to real-world fog. To identify an effective learning rate range that allows the model to overcome the sim-to-real gap, we conducted a hyperparameter sensitivity analysis spanning the following order-of-magnitude scale shifts: 0.01$\times$, 0.1$\times$, 1$\times$ (default), 10$\times$, 20$\times$ and $\geq$ 50$\times$ the default rate. This translates to evaluating learning rates of 0.0002, 0.002, 0.02, 0.2, 0.4 and $\geq$ 1.0. In this analysis, all other training hyperparameters, including batch size, optimiser and weight decay, were kept constant to strictly isolate the impact of gradient scaling on domain adaptation. The models were fined-tuned on the training set of STF-Foggy and evaluated on its test set, with the results plotted in Figure~\ref{fig11} representing the mean performance across three independent random seeds.

As seen in Figure~\ref{fig11}, all evaluated variants (C2F Fixed-Density, C2F Mixed-Density and Multifog KITTI) demonstrate a similar rise and then fall under increasing learning rate scales. Standard fine-tuning protocols (1$\times$ scale, representing $LR=0.02$) result in negative transfer where performance decreases after real-world fine-tuning, resulting in a drop in performance to an average of 0.2450 mAP, 0.2487 mAP and 0.2357 mAP, respectively. However, a tenfold increase in the learning rate (10$\times$ scale, representing $LR=0.2$) provides the strongest performance among the evaluated settings, facilitating stronger adaptation to real-world fog conditions. This leads to a performance peak where all three models exceed the real-only baseline of 0.2553 mAP. Under this 10$\times$ scaling, C2F Fixed-Density achieves 0.2567 mAP, C2F Mixed-Density achieves 0.2610 mAP and Multifog KITTI reaches 0.2670 mAP. This indicates that scaling the learning rate helps mitigate synthetic domain biases, although the magnitude of improvement depends on the synthetic data source.

Pushing the optimisation scale further to a 20$\times$ increase ($LR=0.4$) results in an immediate performance decline, and then this is further exaggerated in training instability and divergence (NaN) at orders of magnitude higher ($\geq$ 50$\times$ scale, representing LR $\geq$ 1.0). On the other hand, lower learning rates (0.1$\times$ and 0.01$\times$) result in severe performance degradation and drop below the real-only baseline by as much as 40.19\%. Ultimately, these findings suggest that while large-scale synthetic datasets provide a highly effective pre-training foundation, synthetic data alone cannot bridge the sim-to-real domain gap. These findings suggest that effective sim-to-real fog adaptation may require synthetic pre-training followed by real-world fine-tuning and learning rate scaling.

\bigskip
\section{Conclusion and Future Work}
\label{sec:conclusion}

\subsection{Conclusion}
In this paper, we introduced Clear2Fog (C2F), a multimodal physics-based pipeline for generating synthetic fog across clear-weather datasets. Beyond providing a scalable fog simulation framework, this work demonstrates that environmental diversity is an important factor alongside dataset scale when constructing synthetic fog datasets. Our data efficiency analysis revealed that, within the evaluated settings, models trained on a 75\% scale mixed-density fog dataset achieved performance with no statistically significant difference compared to models trained on a 100\% fixed-density fog dataset, reducing synthetic training data requirements by 25\% while maintaining detection performance.

We further investigated the role of synthetic fog in sim-to-real transfer. While our experiments showed that synthetic fog alone is not sufficient to fully bridge the domain gap, the results demonstrated its value as a large-scale pre-training foundation. By applying a relative 10x increase in the default fine-tuning learning rate, the pre-trained models adapted more effectively to real-world fog and achieved an absolute increase of up to 0.0117 mAP beyond the real-only baseline. These findings suggest that diverse synthetic fog data can serve as an effective tool for improving object detection in adverse weather when combined with appropriate real-world adaptation. 

\subsection{Limitations}
While the C2F pipeline offers a path forward, its current implementation has specific limitations:
\begin{enumerate}
    \item Real-world validation was restricted to a subset of the Seeing Through Fog dataset totalling to 1,140 foggy images. While the evaluated experiments show consistent trends, evaluating the pipeline on larger, geographically diverse datasets would further solidify its generalisability.
    \item The reliance on monocular depth estimation occasionally results in localised artefacts, especially in scenes with transparent surfaces or high-frequency geometric details. These errors can lead to inconsistent fog placement in specific regions of the frame.
    \item Although the C2F pipeline is an effective tool for pre-training, the computational overhead required to simulate for million-scale datasets is significant. This positions the pipeline primarily as an offline data augmentation tool rather than a real-time training solution.
\end{enumerate}

\subsection{Future Work}
The results of this study suggest several directions for further research:
\begin{enumerate}
    \item Future work could investigate the integration of depth-aware shadow removal as replicating the light diffusion effects of real fog, which naturally softens shadows, may further reduce the domain gap between synthetic and real data.
    \item Evaluating on additional detector families, including transformer-based architectures and newer detectors, would help determine whether the observed data efficiency trends extend beyond the evaluated architectures.
    \item Extending the C2F pipeline to support downstream tasks such as semantic segmentation would provide a more comprehensive assessment of the use of synthetic data across other autonomous perception tasks.
    \item Determining whether synthetic pre-training benefits can be further improved beyond learning rate optimisation
\end{enumerate}

\section*{Acknowledgments}
The authors would like to acknowledge the assistance provided by Research IT and the use of the Barkla High Performance Computing facilities at the University of Liverpool. The authors would also like to thank the participants who took part in the human perceptual study.

\clearpage

\appendix
\section*{Appendix A: Raw Data Efficiency Tables}
\addcontentsline{toc}{section}{Appendix A: Raw Data Efficiency Tables} 
\setcounter{table}{0} 
\renewcommand{\thetable}{A.\arabic{table}}
\label{appendix:A}

\begin{table}[h]
\centering
\caption{Effect of dataset scale on Faster R-CNN model~\cite{r58} performance on the clear validation set. Class-wise metrics represent the mean values across three independent seeds. Average mAP represents the mean of the macro mAP values $\pm$ standard deviation. Relative improvements are based on the 10\% baseline. Numerical values correspond to Figure~\ref{fig10}.}
\label{tabA1}
\setlength{\tabcolsep}{6pt}
\makebox[\textwidth][c]{
    \small
    \begin{tabular}{l c c c c c c}
    \toprule
     & & & & & \multicolumn{2}{c}{Clear Val Set} \\ \cmidrule(lr){6-7}
    \multicolumn{1}{c}{Condition} & \begin{tabular}[c]{@{}c@{}}Dataset\\ Subset\\ (\%)\end{tabular} & \begin{tabular}[c]{@{}c@{}}Vehicle\\ AP\end{tabular} & \begin{tabular}[c]{@{}c@{}}Pedestrian\\ AP\end{tabular} & \begin{tabular}[c]{@{}c@{}}Cyclist\\ AP\end{tabular} & \begin{tabular}[c]{@{}c@{}}Average mAP $\pm$ Std.\\ Dev.\end{tabular} & \begin{tabular}[c]{@{}c@{}}Improvement on\\ 10\% Subset (\%)\end{tabular} \\ \midrule
    \multirow{5}{*}{\begin{tabular}[c]{@{}l@{}}Clear-\\ Weather\\ Baseline\end{tabular}} 
     & 10  & 0.3520 & 0.3033 & 0.1513 & $0.2687 \pm 0.0030$ & 0.00 \\
     & 25  & 0.3877 & 0.3117 & 0.1913 & $0.2969 \pm 0.0017$ & 10.49 \\
     & 50  & 0.4080 & 0.3287 & 0.1897 & $0.3088 \pm 0.0010$ & 14.92 \\
     & 75  & 0.4173 & 0.3453 & 0.2037 & $0.3221 \pm 0.0036$ & 19.87 \\
     & 100 & 0.4260 & 0.3493 & 0.2297 & $0.3350 \pm 0.0026$ & 24.68 \\ \midrule
    \multirow{5}{*}{\begin{tabular}[c]{@{}l@{}}Fixed-\\ Density\\ Fog\end{tabular}} 
     & 10  & 0.3150 & 0.2650 & 0.1227 & $0.2342 \pm 0.0045$ & 0.00 \\
     & 25  & 0.3517 & 0.2643 & 0.1703 & $0.2621 \pm 0.0030$ & 11.91 \\
     & 50  & 0.3737 & 0.2803 & 0.1557 & $0.2699 \pm 0.0010$ & 15.24 \\
     & 75  & 0.3823 & 0.2883 & 0.1763 & $0.2823 \pm 0.0015$ & 20.54 \\
     & 100 & 0.3950 & 0.3003 & 0.1827 & $0.2927 \pm 0.0006$ & 24.98 \\ \midrule
    \multirow{5}{*}{\begin{tabular}[c]{@{}l@{}}Mixed-\\ Density\\ Fog\end{tabular}} 
     & 10  & 0.3247 & 0.2800 & 0.1427 & $0.2491 \pm 0.0030$ & 0.00 \\
     & 25  & 0.3630 & 0.2897 & 0.1620 & $0.2716 \pm 0.0031$ & 9.03 \\
     & 50  & 0.3790 & 0.3037 & 0.1557 & $0.2795 \pm 0.0012$ & 12.20 \\
     & 75  & 0.3930 & 0.3187 & 0.1813 & $0.2977 \pm 0.0012$ & 19.51 \\
     & 100 & 0.4007 & 0.3243 & 0.1987 & $0.3079 \pm 0.0040$ & 23.60 \\ \bottomrule
    \end{tabular}
}
\end{table}

\clearpage

\begin{table}[h]
\centering
\caption{Effect of dataset scale on Faster R-CNN model~\cite{r58} performance on the fixed-density validation set. Class-wise metrics represent the mean values across three independent seeds. Average mAP represents the mean of the macro mAP values ± standard deviation. Relative improvements are based on the 10\% baseline. Numerical values correspond to Figure~\ref{fig10}.}
\label{tabA2}
\setlength{\tabcolsep}{6pt}
\makebox[\textwidth][c]{
    \small
    \begin{tabular}{l c c c c c c}
    \toprule
     & & & & & \multicolumn{2}{c}{Fixed-Density Fog Val Set} \\ \cmidrule(lr){6-7}
    \multicolumn{1}{c}{Condition} & \begin{tabular}[c]{@{}c@{}}Dataset\\ Subset\\ (\%)\end{tabular} & \begin{tabular}[c]{@{}c@{}}Vehicle\\ AP\end{tabular} & \begin{tabular}[c]{@{}c@{}}Pedestrian\\ AP\end{tabular} & \begin{tabular}[c]{@{}c@{}}Cyclist\\ AP\end{tabular} & \begin{tabular}[c]{@{}c@{}}Average mAP $\pm$ Std.\\ Dev.\end{tabular} & \begin{tabular}[c]{@{}c@{}}Improvement on\\ 10\% Subset (\%)\end{tabular} \\ \midrule
    \multirow{5}{*}{\begin{tabular}[c]{@{}l@{}}Clear-\\ Weather\\ Baseline\end{tabular}} 
     & 10  & 0.2650 & 0.2313 & 0.1440 & $0.2134 \pm 0.0142$ & 0.00 \\
     & 25  & 0.2760 & 0.2393 & 0.1553 & $0.2235 \pm 0.0040$ & 4.73 \\
     & 50  & 0.3113 & 0.2540 & 0.1587 & $0.2413 \pm 0.0010$ & 13.07 \\
     & 75  & 0.3200 & 0.2667 & 0.1740 & $0.2536 \pm 0.0025$ & 18.84 \\
     & 100 & 0.3300 & 0.2740 & 0.1867 & $0.2636 \pm 0.0071$ & 23.52 \\ \midrule
    \multirow{5}{*}{\begin{tabular}[c]{@{}l@{}}Fixed-\\ Density\\ Fog\end{tabular}} 
     & 10  & 0.3380 & 0.2980 & 0.1380 & $0.2580 \pm 0.0031$ & 0.00 \\
     & 25  & 0.3623 & 0.2970 & 0.1883 & $0.2825 \pm 0.0042$ & 9.50 \\
     & 50  & 0.3910 & 0.3190 & 0.1740 & $0.2947 \pm 0.0015$ & 14.22 \\
     & 75  & 0.4003 & 0.3333 & 0.2063 & $0.3133 \pm 0.0010$ & 21.43 \\
     & 100 & 0.4090 & 0.3373 & 0.2110 & $0.3191 \pm 0.0029$ & 23.68 \\ \midrule
    \multirow{5}{*}{\begin{tabular}[c]{@{}l@{}}Mixed-\\ Density\\ Fog\end{tabular}} 
     & 10  & 0.3357 & 0.2950 & 0.1530 & $0.2612 \pm 0.0025$ & 0.00 \\
     & 25  & 0.3660 & 0.2990 & 0.1750 & $0.2800 \pm 0.0017$ & 7.20 \\
     & 50  & 0.3870 & 0.3170 & 0.1763 & $0.2934 \pm 0.0012$ & 12.33 \\
     & 75  & 0.3977 & 0.3310 & 0.2030 & $0.3106 \pm 0.0004$ & 18.91 \\
     & 100 & 0.4047 & 0.3353 & 0.2180 & $0.3193 \pm 0.0025$ & 22.24 \\ \bottomrule
    \end{tabular}
}
\end{table}

\clearpage

\begin{table}[h]
\centering
\caption{Effect of dataset scale on Faster R-CNN model~\cite{r58} performance on the mixed-density validation set. Class-wise metrics represent the mean values across three independent seeds. Average mAP represents the mean of the macro mAP values ± standard deviation. Relative improvements are based on the 10\% baseline. Numerical values correspond to Figure~\ref{fig10}.}
\label{tabA3}
\setlength{\tabcolsep}{6pt}
\makebox[\textwidth][c]{
    \small
    \begin{tabular}{l c c c c c c}
    \toprule
     & & & & & \multicolumn{2}{c}{Mixed-Density Fog Val Set} \\ \cmidrule(lr){6-7}
    \multicolumn{1}{c}{Condition} & \begin{tabular}[c]{@{}c@{}}Dataset\\ Subset\\ (\%)\end{tabular} & \begin{tabular}[c]{@{}c@{}}Vehicle\\ AP\end{tabular} & \begin{tabular}[c]{@{}c@{}}Pedestrian\\ AP\end{tabular} & \begin{tabular}[c]{@{}c@{}}Cyclist\\ AP\end{tabular} & \begin{tabular}[c]{@{}c@{}}Average mAP $\pm$ Std.\\ Dev.\end{tabular} & \begin{tabular}[c]{@{}c@{}}Improvement on\\ 10\% Subset (\%)\end{tabular} \\ \midrule
    \multirow{5}{*}{\begin{tabular}[c]{@{}l@{}}Clear-\\ Weather\\ Baseline\end{tabular}} 
     & 10  & 0.2503 & 0.2187 & 0.1390 & $0.2027 \pm 0.0125$ & 0.00 \\
     & 25  & 0.2553 & 0.2200 & 0.1553 & $0.2102 \pm 0.0032$ & 3.70 \\
     & 50  & 0.2943 & 0.2393 & 0.1557 & $0.2298 \pm 0.0021$ & 13.37 \\
     & 75  & 0.2997 & 0.2457 & 0.1750 & $0.2401 \pm 0.0029$ & 18.45 \\
     & 100 & 0.3100 & 0.2540 & 0.1827 & $0.2489 \pm 0.0061$ & 22.79 \\ \midrule
    \multirow{5}{*}{\begin{tabular}[c]{@{}l@{}}Fixed-\\ Density\\ Fog\end{tabular}} 
     & 10  & 0.3190 & 0.2863 & 0.1327 & $0.2460 \pm 0.0040$ & 0.00 \\
     & 25  & 0.3430 & 0.2873 & 0.1760 & $0.2688 \pm 0.0040$ & 9.27 \\
     & 50  & 0.3690 & 0.3033 & 0.1650 & $0.2791 \pm 0.0017$ & 13.46 \\
     & 75  & 0.3797 & 0.3183 & 0.1960 & $0.2980 \pm 0.0017$ & 21.14 \\
     & 100 & 0.3887 & 0.3230 & 0.2020 & $0.3046 \pm 0.0031$ & 23.82 \\ \midrule
    \multirow{5}{*}{\begin{tabular}[c]{@{}l@{}}Mixed-\\ Density\\ Fog\end{tabular}} 
     & 10  & 0.3217 & 0.2897 & 0.1520 & $0.2545 \pm 0.0025$ & 0.00 \\
     & 25  & 0.3557 & 0.2967 & 0.1710 & $0.2745 \pm 0.0015$ & 7.86 \\
     & 50  & 0.3760 & 0.3130 & 0.1767 & $0.2886 \pm 0.0015$ & 13.40 \\
     & 75  & 0.3863 & 0.3260 & 0.1957 & $0.3027 \pm 0.0003$ & 18.94 \\
     & 100 & 0.3933 & 0.3313 & 0.2110 & $0.3119 \pm 0.0017$ & 22.55 \\ \bottomrule
    \end{tabular}
}
\end{table}

\clearpage

\begin{table}[h]
\centering
\caption{Effect of dataset scale on YOLOX-S model~\cite{r60} performance on the clear validation set. Class-wise metrics represent the mean values across three independent seeds. Average mAP represents the mean of the macro mAP values $\pm$ standard deviation. Relative improvements are based on the 10\% baseline. Numerical values correspond to Figure~\ref{fig10}.}
\label{tabA4}
\setlength{\tabcolsep}{6pt}
\makebox[\textwidth][c]{
    \small
    \begin{tabular}{l c c c c c c}
    \toprule
     & & & & & \multicolumn{2}{c}{Clear Val Set} \\ \cmidrule(lr){6-7}
    \multicolumn{1}{c}{Condition} & \begin{tabular}[c]{@{}c@{}}Dataset\\ Subset\\ (\%)\end{tabular} & \begin{tabular}[c]{@{}c@{}}Vehicle\\ AP\end{tabular} & \begin{tabular}[c]{@{}c@{}}Pedestrian\\ AP\end{tabular} & \begin{tabular}[c]{@{}c@{}}Cyclist\\ AP\end{tabular} & \begin{tabular}[c]{@{}c@{}}Average mAP $\pm$ Std.\\ Dev.\end{tabular} & \begin{tabular}[c]{@{}c@{}}Improvement on\\ 10\% Subset (\%)\end{tabular} \\ \midrule
    \multirow{5}{*}{\begin{tabular}[c]{@{}l@{}}Clear-\\ Weather\\ Baseline\end{tabular}} 
     & 10  & 0.3070 & 0.2530 & 0.0947 & $0.2180 \pm 0.0029$ & 0.00 \\
     & 25  & 0.3410 & 0.2537 & 0.1017 & $0.2320 \pm 0.0008$ & 6.37 \\
     & 50  & 0.3613 & 0.2710 & 0.1140 & $0.2487 \pm 0.0029$ & 14.02 \\
     & 75  & 0.3727 & 0.2910 & 0.1263 & $0.2637 \pm 0.0005$ & 20.67 \\
     & 100 & 0.3775 & 0.2945 & 0.1325 & $0.2675 \pm 0.0025$ & 22.91 \\ \midrule
    \multirow{5}{*}{\begin{tabular}[c]{@{}l@{}}Fixed-\\ Density\\ Fog\end{tabular}} 
     & 10  & 0.2733 & 0.2170 & 0.0650 & $0.1851 \pm 0.0016$ & 0.00 \\
     & 25  & 0.3140 & 0.2350 & 0.0947 & $0.2146 \pm 0.0017$ & 15.94 \\
     & 50  & 0.3247 & 0.2523 & 0.1133 & $0.2301 \pm 0.0050$ & 24.31 \\
     & 75  & 0.3400 & 0.2590 & 0.1050 & $0.2347 \pm 0.0007$ & 26.80 \\
     & 100 & 0.3460 & 0.2655 & 0.1200 & $0.2438 \pm 0.0005$ & 31.71 \\ \midrule
    \multirow{5}{*}{\begin{tabular}[c]{@{}l@{}}Mixed-\\ Density\\ Fog\end{tabular}} 
     & 10  & 0.2807 & 0.2330 & 0.0807 & $0.1981 \pm 0.0075$ & 0.00 \\
     & 25  & 0.3117 & 0.2383 & 0.0983 & $0.2161 \pm 0.0010$ & 9.09 \\
     & 50  & 0.3347 & 0.2507 & 0.0940 & $0.2265 \pm 0.0012$ & 14.34 \\
     & 75  & 0.3440 & 0.2710 & 0.1040 & $0.2397 \pm 0.0007$ & 21.00 \\
     & 100 & 0.3510 & 0.2730 & 0.1095 & $0.2445 \pm 0.0015$ & 23.42 \\ \bottomrule
    \end{tabular}
}
\end{table}

\clearpage

\begin{table}[h]
\centering
\caption{Effect of dataset scale on YOLOX-S model~\cite{r60} performance on the fixed-density validation set. Class-wise metrics represent the mean values across three independent seeds. Average mAP represents the mean of the macro mAP values $\pm$ standard deviation. Relative improvements are based on the 10\% baseline. Numerical values correspond to Figure~\ref{fig10}.}
\label{tabA5}
\setlength{\tabcolsep}{6pt}
\makebox[\textwidth][c]{
    \small
    \begin{tabular}{l c c c c c c}
    \toprule
     & & & & & \multicolumn{2}{c}{Fixed-Density Fog Val Set} \\ \cmidrule(lr){6-7}
    \multicolumn{1}{c}{Condition} & \begin{tabular}[c]{@{}c@{}}Dataset\\ Subset\\ (\%)\end{tabular} & \begin{tabular}[c]{@{}c@{}}Vehicle\\ AP\end{tabular} & \begin{tabular}[c]{@{}c@{}}Pedestrian\\ AP\end{tabular} & \begin{tabular}[c]{@{}c@{}}Cyclist\\ AP\end{tabular} & \begin{tabular}[c]{@{}c@{}}Average mAP $\pm$ Std.\\ Dev.\end{tabular} & \begin{tabular}[c]{@{}c@{}}Improvement on\\ 10\% Subset (\%)\end{tabular} \\ \midrule
    \multirow{5}{*}{\begin{tabular}[c]{@{}l@{}}Clear-\\ Weather\\ Baseline\end{tabular}} 
     & 10  & 0.2423 & 0.2190 & 0.0950 & $0.1854 \pm 0.0099$ & 0.00 \\
     & 25  & 0.2777 & 0.2290 & 0.0943 & $0.2003 \pm 0.0021$ & 8.04 \\
     & 50  & 0.2990 & 0.2380 & 0.1087 & $0.2152 \pm 0.0016$ & 16.07 \\
     & 75  & 0.3070 & 0.2517 & 0.1277 & $0.2288 \pm 0.0005$ & 23.41 \\
     & 100 & 0.3115 & 0.2585 & 0.1270 & $0.2323 \pm 0.0035$ & 25.30 \\ \midrule
    \multirow{5}{*}{\begin{tabular}[c]{@{}l@{}}Fixed-\\ Density\\ Fog\end{tabular}} 
     & 10  & 0.3047 & 0.2527 & 0.0850 & $0.2141 \pm 0.0019$ & 0.00 \\
     & 25  & 0.3303 & 0.2577 & 0.0977 & $0.2286 \pm 0.0017$ & 6.77 \\
     & 50  & 0.3433 & 0.2753 & 0.1287 & $0.2491 \pm 0.0024$ & 16.35 \\
     & 75  & 0.3600 & 0.2900 & 0.1230 & $0.2577 \pm 0.0007$ & 20.36 \\
     & 100 & 0.3640 & 0.2945 & 0.1275 & $0.2620 \pm 0.0010$ & 22.37 \\ \midrule
    \multirow{5}{*}{\begin{tabular}[c]{@{}l@{}}Mixed-\\ Density\\ Fog\end{tabular}} 
     & 10  & 0.2973 & 0.2573 & 0.0973 & $0.2173 \pm 0.0019$ & 0.00 \\
     & 25  & 0.3197 & 0.2517 & 0.1073 & $0.2262 \pm 0.0000$ & 4.10 \\
     & 50  & 0.3450 & 0.2710 & 0.1150 & $0.2437 \pm 0.0017$ & 12.15 \\
     & 75  & 0.3460 & 0.2905 & 0.1515 & $0.2627 \pm 0.0065$ & 20.89 \\
     & 100 & 0.3605 & 0.2915 & 0.1265 & $0.2595 \pm 0.0005$ & 19.42 \\ \bottomrule
    \end{tabular}
}
\end{table}

\clearpage

\begin{table}[h]
\centering
\caption{Effect of dataset scale on YOLOX-S model~\cite{r60} performance on the mixed-density validation set. Class-wise metrics represent the mean values across three independent seeds. Average mAP represents the mean of the macro mAP values $\pm$ standard deviation. Relative improvements are based on the 10\% baseline. Numerical values correspond to Figure~\ref{fig10}.}
\label{tabA6}
\setlength{\tabcolsep}{6pt}
\makebox[\textwidth][c]{
    \small
    \begin{tabular}{l c c c c c c}
    \toprule
     & & & & & \multicolumn{2}{c}{Mixed-Density Fog Val Set} \\ \cmidrule(lr){6-7}
    \multicolumn{1}{c}{Condition} & \begin{tabular}[c]{@{}c@{}}Dataset\\ Subset\\ (\%)\end{tabular} & \begin{tabular}[c]{@{}c@{}}Vehicle\\ AP\end{tabular} & \begin{tabular}[c]{@{}c@{}}Pedestrian\\ AP\end{tabular} & \begin{tabular}[c]{@{}c@{}}Cyclist\\ AP\end{tabular} & \begin{tabular}[c]{@{}c@{}}Average mAP $\pm$ Std.\\ Dev.\end{tabular} & \begin{tabular}[c]{@{}c@{}}Improvement on\\ 10\% Subset (\%)\end{tabular} \\ \midrule
    \multirow{5}{*}{\begin{tabular}[c]{@{}l@{}}Clear-\\ Weather\\ Baseline\end{tabular}} 
     & 10  & 0.2280 & 0.2063 & 0.0927 & $0.1757 \pm 0.0109$ & 0.00 \\
     & 25  & 0.2573 & 0.2153 & 0.0940 & $0.1889 \pm 0.0016$ & 7.51 \\
     & 50  & 0.2787 & 0.2250 & 0.1053 & $0.2030 \pm 0.0016$ & 15.54 \\
     & 75  & 0.2870 & 0.2387 & 0.1263 & $0.2173 \pm 0.0008$ & 23.68 \\
     & 100 & 0.2910 & 0.2460 & 0.1260 & $0.2210 \pm 0.0020$ & 25.78 \\ \midrule
    \multirow{5}{*}{\begin{tabular}[c]{@{}l@{}}Fixed-\\ Density\\ Fog\end{tabular}} 
     & 10  & 0.2870 & 0.2440 & 0.0823 & $0.2044 \pm 0.0019$ & 0.00 \\
     & 25  & 0.3140 & 0.2517 & 0.1020 & $0.2226 \pm 0.0017$ & 8.90 \\
     & 50  & 0.3263 & 0.2680 & 0.1263 & $0.2402 \pm 0.0037$ & 17.51 \\
     & 75  & 0.3410 & 0.2820 & 0.1210 & $0.2480 \pm 0.0000$ & 21.33 \\
     & 100 & 0.3455 & 0.2850 & 0.1240 & $0.2515 \pm 0.0015$ & 23.04 \\ \midrule
    \multirow{5}{*}{\begin{tabular}[c]{@{}l@{}}Mixed-\\ Density\\ Fog\end{tabular}} 
     & 10  & 0.2833 & 0.2513 & 0.0967 & $0.2104 \pm 0.0026$ & 0.00 \\
     & 25  & 0.3113 & 0.2507 & 0.1100 & $0.2240 \pm 0.0005$ & 6.46 \\
     & 50  & 0.3350 & 0.2690 & 0.1165 & $0.2402 \pm 0.0005$ & 14.16 \\
     & 75  & 0.3460 & 0.2850 & 0.1240 & $0.2517 \pm 0.0007$ & 19.63 \\
     & 100 & 0.3513 & 0.2887 & 0.1263 & $0.2554 \pm 0.0008$ & 21.39 \\ \bottomrule
    \end{tabular}
}
\end{table}

\clearpage

\printbibliography

\end{document}